\documentclass[10pt,twocolumn,letterpaper]{article}

\usepackage[pagenumbers]{cvpr} 

\usepackage{graphicx}
\usepackage{amsmath}
\usepackage{amssymb}
\usepackage{booktabs}
\usepackage{algorithm}
\usepackage{algorithmic}
\usepackage{comment}
\usepackage{dsfont}
\usepackage{xspace}
\usepackage{xcolor}
\usepackage{enumerate}
\usepackage{multirow}
\usepackage[accsupp]{axessibility}  

\usepackage[pagebackref,breaklinks,colorlinks]{hyperref}

\usepackage[capitalize]{cleveref}
\crefname{section}{Sec.}{Secs.}
\Crefname{section}{Section}{Sections}
\Crefname{table}{Table}{Tables}
\crefname{table}{Tab.}{Tabs.}

\def\cvprPaperID{779} 
\def\confName{CVPR}
\def\confYear{2023}

\newcommand{\blue}[1]{\textcolor{blue}{#1}}
\newcommand{\OURS}{PoseExaminer\xspace}

\newcommand{\de}[1]{{\small{\textcolor{blue}{(#1)}}}}

\newcommand{\red}[1]{\textcolor{red}{#1}}

\begin{document}

\title{\OURS: Automated Testing of Out-of-Distribution Robustness \\in Human Pose and Shape Estimation}


\author{
Qihao Liu$^1$ \qquad Adam Kortylewski$^{2,3}$ \qquad Alan Yuille$^1$ \\
{\normalsize $^1$Johns Hopkins University \quad $^2$Max Planck Institute for Informatics \quad $^3$University of Freiburg}
}
\maketitle

\begin{abstract}
Human pose and shape (HPS) estimation methods achieve remarkable results.
However, current HPS benchmarks are mostly designed to test models in scenarios that are similar to the training data.
This can lead to critical situations in real-world applications when the observed data differs significantly from the training data and hence is out-of-distribution (OOD).
It is therefore important to test and improve the OOD robustness of HPS methods.
To address this fundamental problem, we develop a simulator that can be controlled in a fine-grained manner using interpretable parameters to explore the manifold of images of human pose, e.g. by varying poses, shapes, and clothes.
We introduce a learning-based testing method, termed \OURS, that automatically diagnoses HPS algorithms by searching over the parameter space of human pose images to find the failure modes.
Our strategy for exploring this high-dimensional parameter space is a multi-agent reinforcement learning system, in which the agents collaborate to explore different parts of the parameter space.
We show that our \OURS discovers a variety of limitations in current state-of-the-art models that are relevant in real-world scenarios but are missed by current benchmarks.
For example, it finds large regions of realistic human poses that are not predicted correctly, as well as reduced performance for humans with skinny and corpulent body shapes. 
In addition, we show that fine-tuning HPS methods by exploiting the failure modes found by \OURS improve their robustness and even their performance on standard benchmarks by a significant margin.
The code are available for research purposes at \href{https://github.com/qihao067/PoseExaminer}{https://github.com/qihao067/PoseExaminer}.

\end{abstract}

\section{Introduction}
\label{sec:intro}

\begin{figure}
    \centering
    \includegraphics[width=\linewidth]{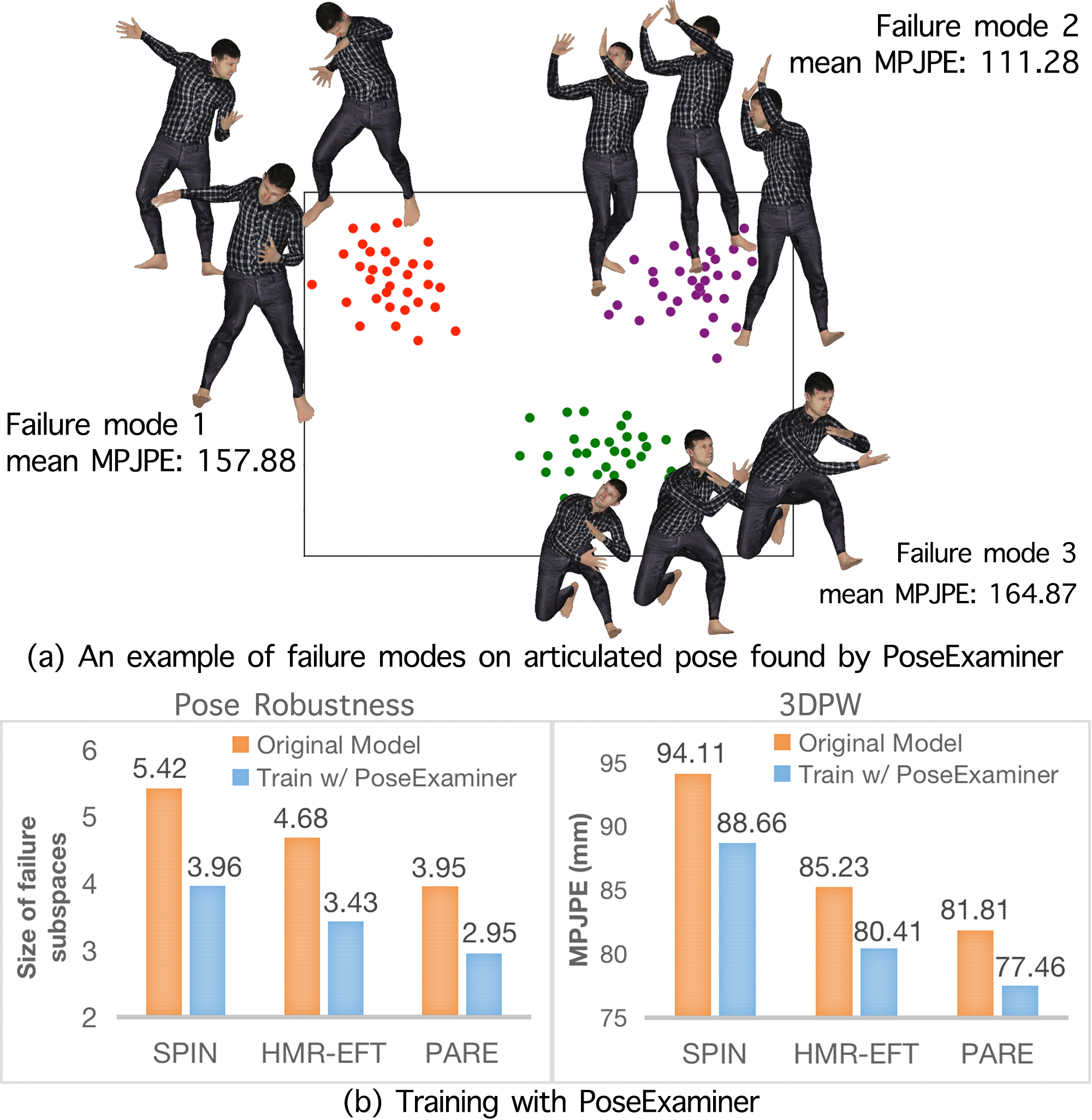}    \caption{\OURS is an automatic testing tool used to study the performance and robustness of HPS methods in terms of articulated pose, shape, global rotation, occlusion, etc. It systematically explores the parameter space and discovers a variety of failure modes. (a) illustrates three failure modes in PARE~\cite{kocabas2021pare} and (b) shows the efficacy of training with \OURS.}
    \vspace{-1.0em}
    \label{fig:teaser}
\end{figure}

In recent years, the computer vision community has made significant advances in 3D human pose and shape (HPS) estimation.
But despite the high performance on standard benchmarks, current methods fail to give reliable predictions for configurations that have not been trained on or in difficult viewing conditions such as when humans are significantly occluded~\cite{kocabas2021pare} or have unusual poses or clothing ~\cite{patel2021agora}.  
This lack of robustness in such out-of-distribution (OOD) situations, which typically would not fool a human observer, is generally acknowledged and is a fundamental open problem for HPS methods that should be addressed.

The main obstacle, however, to testing the robustness of HPS methods is that test data is limited, because it is expensive to collect and annotate.
One way to address this problem is to diagnose HPS systems by generating large synthetic datasets that randomly generate images of humans in varying poses, clothing, and background~\cite{patel2021agora, cai2021playing}.
These studies suggest that the gap between synthetic and real domains in HPS is sufficiently small for testing on synthetic to be a reasonable strategy, as we confirm in our experiments.

These methods, however, only measure the \textit{average performance} on the space of images. In sensitive domains that use HPS (\eg autonomous driving), the \textit{failure modes}, where performance is low, can matter more. Most importantly, despite large scale, these datasets lack diversity in some dimensions. For example, they mostly study pose for common actions such as standing, sitting, walking, etc.

Inspired by the literature on adversarial machine learning~\cite{carlini2017towards,explaining_adv}, recent approaches aim to systematically explore the parameter space of simulators for weaknesses in the model performance. 
This has proven to be an effective approach for diagnosing limitations in image classification~\cite{shu2020identifying}, face recognition~\cite{ruiz2022simulated}, and path planning~\cite{rempe2022generating}. 
However, they are not directly applicable to testing the robustness of the high-dimensional regression task of pose estimation. For example,~\cite{ruiz2022simulated,shu2020identifying} were designed for simple binary classification tasks and can only optimize a limited number of parameters, and ~\cite{rempe2022generating} only perturbs an initial real-world scene to generate adversarial examples, which does not enable the full exploration of the parameter space.  

In this work, we introduce \OURS (Fig.~\ref{fig:teaser}), a learning-based algorithm to automatically diagnose the robustness of HPS methods.
It efficiently searches through the high-dimensional continuous space of a simulator of human images to find failure modes.
The strategy in \OURS is a multi-agent reinforcement learning approach, in which the agents collaborate with one another to search the latent parameter space for model weaknesses.
More specifically, each agent starts at different random initialized seeds and explores the latent parameter space to find failure cases, while at the same time avoiding exploring the regions close to the other agents in the latent space. 
This strategy enables a highly parallelizable way of exploring the high-dimensional continuous latent space of the simulator. 
After converging to a local optimum, each agent explores the local parameter space to find a connected failure region that defines a whole subspace of images where the pose is incorrectly predicted  (\ie failure mode). 
We demonstrate that very large subspaces of failures exist even in the best HPS models.
We use the size  of these failure subspaces together with the success rate of the agents as a new measure of out-of-distribution robustness.

Our experiments on four state-of-the-art HPS models show that \OURS successfully discovers a variety of failure modes that provide new insights about their real-world performance.
For example, it finds large subspaces of realistic human poses that are not predicted correctly, and reduced performance for humans with skinny and corpulent body shapes. 
Notably, we find that the failure modes found in synthetic data generalize well to real images:
In addition to using \OURS as a new benchmark, we also find that fine-tuning SOTA methods using the failure modes discovered by \OURS enhances their robustness and even the performance on 3DPW~\cite{pw3d} and AIST++~\cite{aist,aist++} benchmarks. 
We also note that computer graphics rendering pipelines become increasingly realistic, which will directly benefit the quality of our automated testing approach.

In short, our contributions are three-fold:
\begin{itemize}
\item We propose \OURS, a learning-based algorithm to automatically diagnose the robustness of human pose and shape estimation methods.
Compared to prior work, \OURS is the first to efficiently search for a variety of failure modes in the high-dimensional continuous parameter space of a simulator using a multi-agent reinforcement learning framework.

\item We introduce new metrics, which have not been possible to measure before, for quantifying the robustness of HPS methods based on our automated testing framework. We perform an in-depth analysis of current SOTA methods, revealing a variety of diverse failure modes that generalize well to real images.

\item We show that the failure modes discovered by \OURS can be used to significantly improve the real-world performance and robustness of current methods. 

\end{itemize}
\section{Related work}
\label{sec:relatedwork}

\noindent\textbf{Human Pose and Shape Estimation.} Current methods typically follow one of two paradigms: 
\textit{Regression-based methods}~\cite{guler_2019_CVPR,kanazawa_hmr,omran2018nbf,pavlakos2018humanshape,Tan,tung2017self,kolotouros2019learning} directly estimate 3D human parameters from RGB images. However, due to the difficulty of collecting highly accurate 3D ground truth, there aren't enough labeled training examples. One solution is to use weak supervision such as 2D keypoints, silhouettes, and body part segmentation in the training either as auxiliary consistency loss~\cite{kanazawa_hmr,pavlakos2018humanshape,Tan,zanfir2020weakly,tung2017self} or intermediate representations~\cite{pavlakos2018humanshape,omran2018nbf,alldieck2019learning}. In addition, pseudo-groundtruth~\cite{joo2021exemplar} labels are demonstrated to be effective in improving the generalization of regression-based methods like HMR~\cite{kanazawa_hmr}. 
\textit{Optimization-based methods}~\cite{SMPL-X:2019,bogo2016keep,xiang2019monocular} require parametric human models like SMPL~\cite{scape,looper_smpl,SMPL-X:2019}.
SMPLify~\cite{bogo2016keep} is the first automated method to fit 2D keypoint detection to the SMPL model. After that, a lot of work has been done to utilize more information during the fitting procedure~\cite{lassner_up3d, huang_mvsmplify, totalcapture}. Also, the optimization-based model can be used to provide better supervision for training: SPIN~\cite{SPIN:ICCV:2019} combines HMR~\cite{kanazawa_hmr} with SMPLify~\cite{bogo2016keep} in a training loop. At each step, HMR initializes SMPLify and then the latter provides better supervision for the former.

These methods make great success on standard benchmarks, but the difficulty in obtaining annotations also causes limited test sets with small diversity, making it difficult to evaluate how close the field is to fully robust and general solutions~\cite{patel2021agora}. Recognizing this issue, recent work~\cite{patel2021agora,cai2021playing} builds synthetic datasets with various subjects for training and testing, yielding promising results. However, these fixed datasets mainly focus on common poses like sitting and standing, and they only care about the average performance, while we consider the failure modes under a large variety of poses. More importantly, \OURS takes a big step forward and automatically diagnoses the performance and robustness in a highly efficient way.

\noindent\textbf{Testing CV with Synthetic Data.} Although there is a relative paucity of work in this area, testing computer vision models with synthetic data is not a novel idea~\cite{pinto2008establishing, mayer2016large, johnson2017clevr, kortylewski2018empirically, kortylewski2019analyzing, ruiz2020morphgan, qiu2017unrealcv, zhang2018unrealstereo}. Previous work mainly focuses on training CV models with synthetic images ~\cite{virtualkitti,synthia,playing_for_data,dosovitskiy2017carla,kortylewski2018training,gecer2018semi,gecer2020synthesizing,marriott20213d}. In contrast to this body of work, we propose to search the parameter space of a human simulator in order to test the HPS model in an adversarial manner. 
However, different from the traditional adversarial attacks~\cite{szegedy2013intriguing, papernot2017practical, carlini2017towards, explaining_adv, madry2018towards, ruiz2020disrupting}, we focus on finding OOD examples that are realistic and in some cases, common in real life, but still fool the models. More importantly, our method can find not only a single isolated adversarial example, but also subspaces of parameters that lie in the latent space of a human simulator.

Recently, there is very interesting work that systematically explores the
parameter space of simulators for weaknesses in the model
performance~\cite{shu2020identifying,ruiz2022simulated,rempe2022generating}. However, ~\cite{rempe2022generating} only perturbs an initial real-world scene to generate adversarial examples, which does not enable the full exploration of the parameter space, and ~\cite{ruiz2022simulated,shu2020identifying} are designed for simple binary classification tasks and only optimize a limited number of parameters. We focus on human pose and shape, a more difficult high-dimensional regression task. More importantly, in comparison to ~\cite{ruiz2022simulated,shu2020identifying}, we simultaneously search the entire parameter space for as many various failure modes as possible and determine the exact boundary of them, and we also demonstrate that the failure modes discovered by \OURS are beneficial to solving real-world problems by revealing the limitations of current approaches and improving their robustness and performance.

\begin{figure}
    \centering
    \includegraphics[width=\linewidth]{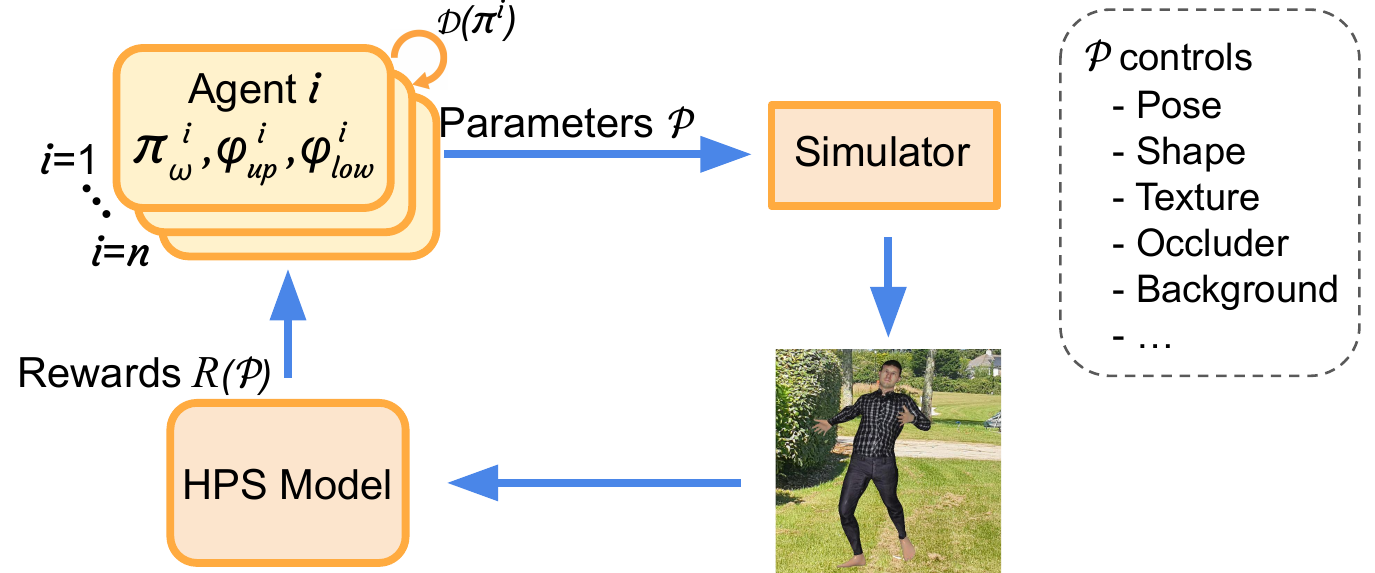}
    \caption{\textbf{\OURS model pipeline.} Multiple RL agents collaborate to  search for worst cases generated by a policy $\pi_{\omega}$ and find subspaces boundaries defined by  $\phi_{up}$ and $\phi_{low}$. The human simulator is conditioned on parameters generated by the agents. $m$ images are then generated for each agent and are used to test a given HPS method. The prediction error of the HPS model serves as the reward signal to update the policy parameters of agents.}
    \vspace{-1.0em}
    \label{fig:pipeline}
\end{figure}

\section{Adversarial Examiner for HPS} \label{sec:method}
\OURS aims to search the entire parameter space for as many different failure modes as possible, and then determine the boundaries of all failure subspaces.
For our purpose, we first need to define the parameter spaces $\mathcal{P}$ of a simulator to control different features of human images (Sec.~\ref{sec:ProDef}). 
Then we need an efficient policy to search over the parameter spaces for a variety of failure modes. To do so, we divide the search process into two phases: We first localize as many different failure cases as possible (Sec.~\ref{sec:phase1}), and then find the boundaries for all failure cases to form subspaces (Sec.~\ref{sec:phase2}). An illustration is provided in Fig.~\ref{fig:pipeline}.

\subsection{Human Image Generation}
\label{sec:ProDef}
To represent human shape and pose, we use SMPL~\cite{loper2015smpl} parameters. SMPL represents the body pose and shape by $\Theta$, which consists of the pose $\theta\in \mathbb{R}^{72}$ and shape $\beta\in\mathbb{R}^{10}$ parameters. Here we only consider the pose of human body, thus we omit the last two joints (\ie the hands). To better test the HPS methods, we divide the rest of the pose parameters into two parts: global rotation $\theta_g\in \mathbb{R}^{3}$ and the articulated pose $\theta_a\in \mathbb{R}^{63}$. The former represents the pose vector of the pelvis and the latter represents the vector of the remaining 21 joints. Given these parameters, the SMPL model is a differentiable function that outputs a posed 3D mesh $\mathcal{M}(\theta,\beta)\in\mathbb{R}^{6890\times 3}$. We use PyTorch3D to render 3D human mesh. During rendering, we also consider the skin color $c_{skin}\in\mathbb{R}^{3}$ which consists of the RGB value(scale) of each vertex, and clothing $\psi_{text}$ that is used to generate UV maps to texture the mesh surface. 

In addition to the parameters related to humans, we also consider the background $\psi_{bg}$ which controls the background images, the occluder $\{\psi_{occ},t_{occ}\}$, in which $\psi_{occ}$ controls the shape and texture and $t_{occ}\in\mathbb{R}^3$ consists of the position $(u_{occ},v_{occ})$ and rotation angle $r_{occ}$ of the patch. Finally, we also consider the lighting $l\in\mathbb{R}^4$ of which the elements control the $xyz$ position and the intensity.

Note that our preliminary experiments suggest that common and noiseless backgrounds, clothing, and textures of occluders are not the key factors affecting performance compared to other factors (see Supp.), and generating realistic images of them via a generative model with parameters $\psi_{(\cdot)}$ is time-consuming. Therefore, following PARE~\cite{kocabas2021pare} that directly uses occluding patches, we collect real images including 150 backgrounds of 10 scenarios, 40 high-quality UV maps, and 12 occluders and use $\psi_{(\cdot)}$ as a dictionary.

In this work, we mainly focus on single-person HPS estimation, and because most methods require cropped images with the target person in the center, we directly use a perspective camera with focal lengths 60 mm and fixed scale and translation parameters. We can study the viewpoints by changing the global rotation $\theta_g$.

\subsection{Finding the Worst-case Poses}
\label{sec:phase1}
Now we have defined all the parameters required to generate images, we want to efficiently search the space, finding parameters such that the samples generated with them fool a given algorithm and the errors are above the threshold $T$. We take the articulated pose $\theta_a$ as an example in the following two subsections. Other parameters are much easier to study. We want to emphasize that our solution is general and can handle all objectives mentioned above. We analyze all parameters in the experiments section.

\OURS has no knowledge of the given HPS method $H$ and the simulator $S$ is non-differentiable, so we use reinforcement learning (RL) to search for the adversarial parameters. We adopt a policy gradient method. To efficiently handle the high-dimensional pose space and learn as many different failure modes as possible, we employ $n$ RL agents to search cooperatively. Another challenge is to ensure the poses we find are physically possible. Thus, we use VPoser~\cite{SMPL-X:2019}, a VAE-based pose prior, to constrain the pose and let the agents search the 32-dimensional latent space in this phase. Other advanced priors can be used here. The prior allows our model to start from different simple poses and learn valid and relatively common adversarial examples. The limitation of using a prior that trained on a fixed dataset such as AMASS~\cite{mahmood2019amass} is solved in phase 2 (Sec.~\ref{sec:phase2}).

Specifically, for agent $i$, we define a policy $\pi_{\omega^i}$ parameterized by $\omega$ that we can sample parameters $z^i\sim \pi_{\omega^i}(z^i)$. Then we use the pre-trained decoder $f$ of VPoser to get the articulated pose $\theta^i_a=f(z^i)$, and render an image $I^i$ via simulator $S$, $I^i=S(\theta^i_a, \Psi^i)$, where $\Psi^i$ denotes the other controllable parameters in Sec.~\ref{sec:ProDef}. After that we can test the method $H$ on $I^i$ and get the error. To avoid the ambiguity along depth, we use 2D MPJPE $err^i_{2D}$ so that the agents only focus on finding poses that cause large error even after being projected to 2D. We define a reward $R(z^i)=c-err^i_{2D}$ where $c$ is a constant. Then we can optimize the parameters $\omega^i$ by maximizing
\begin{align}
    L(\omega^i)=\mathbb{E}_{z^i\sim \pi_{\omega^i}}[R(z^i)] + \mathds{1}_{\{i\neq b\}}\gamma\mathbb{E}[D(\pi_{\omega^i},\pi_{\omega^b})]
\label{eq1}
\end{align}
where $\mathds{1}_{\{\cdot\}}$ denotes the indicator function, $\gamma (>0)$ denotes a weighting factor, and $D(\pi_{\omega},\pi_{\omega'})$ is the distance measure between two policies $\pi_{\omega}\text{ and }\pi_{\omega'}$ (here we use the $L2$ distance between $z\text{ and }z'$).

The gradient of the second term in Eq.~\ref{eq1} is easy to obtain by computing the mean distance between agents. Gradient of the first term is computed following the REINFORCE rule
\begin{align}
    \nabla_{\omega^i}J(\omega^i)=\mathbb{E}_{z^i\sim \pi_{\omega^i}(\cdot)}[\nabla_{\omega^i}\log(\pi_{\omega^i})R(z^i)]
\label{eq2}
\end{align}
and an unbiased, empirical estimate of Eq.~\ref{eq2} is:
\begin{align}
    \nabla_{\omega^i}\hat{J}(\omega^i)=\frac{1}{K}\sum_{k=1}^K\nabla_{\omega^i}\log(\pi_{\omega^i})(R(z^i)-b)
\label{eq3}
\end{align}
in which $K$ is the number of parameters sampled and $b$ is a baseline and is updated every iteration by $b=(1-\tau)b+\tau R(z^i)$. 
Please see Supp. for the pseudo-code.

\subsection{Finding Failure Subspaces}
\label{sec:phase2}
After learning multiple failure cases in phase 1, we search the space around these points for failure subspaces. However, VPoser has several limitations. For example, the distances between individual poses are not preserved in the VAE latent space, implying that the space is not necessarily continuous or smooth mathematically. More importantly, it is trained on a finite dataset, resulting in a biased distribution. Therefore, in this phase, we directly search the pose space using a hierarchical searching policy. We also introduce two new metrics to evaluate the subspaces and boundaries we discover (Sec.~\ref{sec:DiaMet}), and demonstrate the effectiveness of our searching policy.

Assume agent $i$ has discovered an adversarial point $z^i$, then we have the corresponding adversarial pose $\theta^i_a=f(z^i), \theta^i_a\in \mathbb{R}^{63}$. We define two non-negative variables $\phi^i_{up}\in \mathbb{R}^{63}$ and $\phi^i_{low}\in \mathbb{R}^{63}$ that denote the upper bound and the lower bound respectively. To ensure searching efficiency and accuracy, for each iteration, we select one joint, \eg joint $j$, according to the kinematic tree, from parent to child nodes, and only consider the space around the upper bound or the lower bound of the three rotation directions of the selected joint (\ie 3 out of $63\times 2$ dimensions). 
Given a step size $\delta \in \mathbb{R}^{63}$, we can sample $m$ poses $\theta^i_{a,j}$ either from $\mathcal{U}(\theta^i_a+\phi^i_{up},\theta^i_a+\phi^i_{up}+\delta)$ to optimize upper bound, or from $\mathcal{U}(\theta^i_a-\phi^i_{low}-\delta,\theta^i_a-\phi^i_{low})$ to optimize lower bound. The subscript $j$ means only the elements corresponding to joint $j$ are different in these samples. After that, we test the method $H$ and compute the errors. To ensure the poses are physically possible, we consider the limits of joint angles and use Bounding Volume Hierarchies~\cite{teschner2005collision} to avoid self-collisions. We update the boundary when the minimum error of all samples is higher than the adversarial threshold $T$ and the pose is valid. Note that the latter is computed at joint level. For computational efficiency, we use a highly parallelized implementation of BVH following VPoser~\cite{SMPL-X:2019}. The step size decreases linearly w.r.t the minimum error. Pseudo-code and details of this phase are provided in Supp.

\section{Benchmarking and Improving Robustness with \OURS}
\label{sec:advexaminer}
\OURS contributes to the field of HPS in two aspects: (1) It provides a highly efficient way to systematically study the performance, robustness, and weaknesses of a given method (Sec.~\ref{sec:DiaMet}). (2) It can be used to improve the robustness and performance of existing methods (Sec.~\ref{sec:impRob}).

\subsection{Quantifying Robustness with \OURS}
\label{sec:DiaMet}
Using \OURS to diagnose HPS methods is intuitive. All parameters in Sec.~\ref{sec:ProDef} can be studied separately or jointly by optimizing one or more types of parameters while fixing the others. Note that by using different sets of hyperparameters of \OURS, such as the searching boundary of policy $\pi_{\omega}$, the adversarial threshold $T$, the limits of joint angles, and so on, we can get examiners with different difficulty levels. For a fair comparison, we select one set of hyperparameters for all benchmark experiments. Please refer to the Supp. for details.

\noindent\textbf{Metrics.} We introduce several new metrics in this work. Two of them are used to assess the quality of failure modes and their boundaries learned by \OURS, and the others are used to analyze the robustness of HPS methods.

To assess the quality of boundaries, we sample $n\times 200$ examples from the failure subspaces and test the given method on these examples ($n$ is the number of agents).
\textbf{Percentage of non-adversarial examples (Pnae)} represents the percentage of examples with 3D MPJPE smaller than the adversarial threshold $T$. We also consider \textbf{minimum of the mean per joint position error over failure modes (minMPJPE)}. Usually minMPJPE is slightly smaller than the adversarial threshold T. Pnae ensures that the boundaries are `small' enough and no non-adversarial examples are included, while minMPJPE ensures that the boundaries are `large/accurate' enough to be close to the threshold. 

To evaluate the robustness of a given method, we consider the failure modes \OURS learns. \textbf{Region Size} denotes the $L2$ distance between the upper and lower bounds of the subspaces, \ie $\frac{1}{n}\sum_{i=1}^n||\phi^i_{up}+\phi^i_{low}||_2$ (here $\phi^i_{up}$ and $\phi^i_{low}$ are non-negative). \textbf{Success Rate} refers to the percentage of agents that find failure modes in specific time steps. We also consider the \textbf{mean MPJPE over failure modes (meanMPJPE)} and the \textbf{maximum MPJPE over failure modes (maxMPJPE)} that describe the average and worst performance under failure modes.
A method is more robust if these metrics are lower.

\subsection{Improving Robustness Using Failure Subspaces}
\label{sec:impRob}
With the failure modes being discovered, the method can be improved by fine-tuning it using the detected modes. An intuitive way is to directly sample poses from the learned subspaces and add them to training data. However, as mentioned previously, current in-distribution (ID) datasets are limited and only cover a small space, while \OURS searches a large space and finds many failure modes with different types and difficulty levels. Due to model capacity and the forgetting problem, directly using a strict examiner could harm the performance on standard benchmarks and simple poses (see Sec.~\ref{sec:ablStudy}). To make the fine-tuning procedure stable and achieve good results on both ID and OOD datasets, we use curriculum learning and provide an efficient way to use \OURS to improve algorithms.

Specifically, we fine-tune the model following an easy-to-difficult schedule, using a pre-defined ordered list of hyperparameters that initializes examiners taking into account different factors and difficulty levels. For each loop, after finding multiple failure modes, we sample $m$ cases per failure mode and add them to an adversary set $\mathcal{F}$, then we fine-tune the given method on a weighted combination of data from the original training set $\mathcal{T}$ and the adversary set $\mathcal{F}$. For each loop, $\epsilon$ percent of data are sampled from $\mathcal{F}$ and $(1-\epsilon)$ percent of data are from $\mathcal{T}$. Pseudo-code of fine-tuning with \OURS is provided in the Supp.

We apply a discount factor $\alpha$ to the learning rate during fine-tuning. Since we render all images with a fixed camera, we remove the loss terms of 2D keypoints and camera parameters and only consider the terms directly related to the SMPL parameters and 3D keypoints. For other details of the hyperparameters, data augmentations, and training procedures, we strictly follow the original methods.

\section{Experiments}
\label{sec:Exp}
We then highlight the new capabilities our \OURS enables. 
Sec.~\ref{sec:perGap} shows that the failure modes discovered by \OURS generalize well to real images.
Then we use our method to diagnose four HPS methods and provide in-depth analyses in Sec.~\ref{sec:poseRobustness} and~\ref{sec:otherRobustness}.
The failure modes are then used to improve the current arts in Sec.~\ref{sec:train} and~\ref{sec:ablStudy}.

\noindent\textbf{Implementation Details.}
We use a multi-variable Gaussian policy $\pi_{\omega}(\cdot)=\mathcal{N}(\mu_\pi,\sigma^2_\pi)$ in phase one. The variance is fixed $\sigma^2_\pi=0.05\times I$ and the mean $\mu_\pi$ is learned by a single layer perceptron. We set $T=90$ as the adversarial threshold. $c=50$ and $\gamma=0.2$. In the method analyses in Sec.~\ref{sec:poseRobustness} and~\ref{sec:otherRobustness}, we employ $n=320$ agents and each agent samples $m=50$ failure cases in each iteration during training, and $m=200$ to compute the metrics in Sec~\ref{sec:DiaMet}. In the training experiments in Sec.~\ref{sec:train} and~\ref{sec:ablStudy}, we set $n=40$ and $m=500$ to generate the adversary set $\mathcal{F}$. Adam optimizer with a learning rate of $0.2$ is used. Each agent is trained for at most 300 iterations in phase one and 600 in total.

\noindent\textbf{Models.}
We consider four HPS methods: HMR~\cite{kanazawa_hmr}, SPIN~\cite{SPIN:ICCV:2019}, HMR-EFT~\cite{joo2021exemplar}, and PARE~\cite{kocabas2021pare}. In the method analyses in Sec.~\ref{sec:poseRobustness} and~\ref{sec:otherRobustness}, we directly use the official pre-trained model and/or PyTorch-based implementations provided in MMHuman3D~\cite{mmhuman3d}. In the training experiments in Sec.~\ref{sec:train} and~\ref{sec:ablStudy}, we fine-tune these methods strictly following their original implementation details, with a discount factor $\alpha=0.05$ applied to the original learning rate. We set sample rate $\epsilon=0.1$, \ie $10\%$ of the training data is from the adversary set $\mathcal{F}$ and $90\%$ is from the original training sets.

\noindent\textbf{Datasets.} 
The 3DPW test split~\cite{pw3d} and AIST++~\cite{aist,aist++} are used for evaluation. 3DPW contains subjects performing common actions such as sitting, walking, etc, and is thus classified as an ID dataset. AIST++ is a more complex dataset in which professional dancers dance in front of a white background, surrounded by nine cameras. It contains a large variety of poses and is used as an ODD dataset. We convert the videos to images with 60 FPS following~\cite{aist++}.

A disadvantage of the AIST++ dataset is that it only contains pseudo labels, and some have completely incorrect annotations. We filter out these images by ensuring  (1) the consistency of the SMPL, 3D keypoints, and 2D keypoints annotations, (2) the smoothness of the depth estimates, and (3) the smoothness of the 3D joints on the face. Then we manually check 1000 randomly selected images to ensure the quality of the annotations. (Please see Supp. for details)

The remaining annotations are sufficiently accurate and can be used for evaluation because they are regressed from images captured by nine cameras surrounding the subjects. We randomly select 40K images from the clean AIST++ images and named them cAIST. Also, we use the Vposer encoder to obtain the latent distribution of each pose in the clean AIST++ images, select poses that have more than 15 dimensions latent parameters larger than 1, and randomly select 40K poses from them to form a subset named cAIST-ext, which mainly contains extreme and OOD poses.

\noindent\textbf{Evaluation Metrics.} In addition to the new metrics provided in Sec~\ref{sec:DiaMet}, we also report Procrustes-aligned mean per joint position error (PA-MPJPE) and mean per joint position error (MPJPE) in $mm$. We only report per vertex error (PVE) for 3DPW because AIST++ does not provide shape annotations and it uses mean shape instead.

\subsection{Generalization from Synthetic to Real}
\label{sec:perGap}
\begin{table}
\renewcommand\arraystretch{0.9}
    \centering
    \setlength\tabcolsep{3pt}
    \resizebox{1\columnwidth}{!}{
        \begin{tabular}{l|ccc|ccc}
        \toprule
        &\multicolumn{3}{c}{real 3DPW}& \multicolumn{3}{|c}{sync 3DPW}\\
        \cmidrule(lr){2-4} \cmidrule(l){5-7}
        & {\small MPJPE} & {\small PA-MPJPE}& {\small PVE} & {\small MPJPE} & {\small PA-MPJPE} &{\small PVE } \\
        \midrule
        SPIN~\cite{SPIN:ICCV:2019} & 96.9 & 59.2 & 135.1  & 93.9 {\footnotesize(-3.0)} & 60.9 {\footnotesize(+1.7)}  & 130.9 {\footnotesize(-4.2)}   \\
        PARE~\cite{kocabas2021pare} & 82.0 & 50.9 & 97.9 & 79.6 {\footnotesize(-2.4)} & 51.7 {\footnotesize(+0.8)} & 94.5 {\footnotesize(-3.4)} \\
        PARE{\small (w. 3DPW)}~\cite{kocabas2021pare} & 74.5 & 46.5 & 88.6 & 75.8 {\footnotesize(+1.3)} & 49.4 {\footnotesize(+2.9)} & 89.5 {\footnotesize(+0.9)} \\
        \bottomrule
        \end{tabular}}
    \caption{\textbf{Measuring the performance gap of SPIN and PARE on real and synthetic images.} We can observe that both models perform very well on synthetic data, which has exactly the same poses as the real dataset, indicating that the domain gap is small and even favorable towards the synthetic data.}
    \vspace{-1.0em}
    \label{table:realSyn}
\end{table}
In this section, we demonstrate that (1) with careful design, the performance gap between real and synthetic is small on both ID and OOD datasets, proving that evaluation on synthetic is worthwhile, and (2) the failure modes discovered by \OURS generalize well to real images.

\noindent\textbf{Performance Gap between Real and Synthetic.}
To generate synthetic versions of the 3DPW and cAIST-EXT datasets, we need to obtain the original background image, and then render ground-truth human meshes on top of it. We first use a video instance segmentation method~\cite{wu2022defense} to remove the people from the image, and then use image imprinting~\cite{suvorov2022resolution} to fill in the gaps left by the removal of the people. After that,  for each video, we select UV textures that look the most like the removed people in order to render the synthetic human onto the background image. We also consider the overlapped masks when running the segmentation method, and add an occluding patch that looks similar to the original occluder in the image to restore the occlusions. (Please see the Supp. for figures of each step.)

We evaluate the official pretrained models of SPIN and PARE on both synthetic and real versions of 3DPW and cAIST-EXT datasets and provide results in Tab.~\ref{table:realSyn} and Supp. We can see that all models achieve similar performance on the corresponding synthetic and real versions, no matter on the ID or OOD dataset, or on common (simple) or extreme (complex) poses. PVE on synthetic 3DPW is much smaller than on real. We think this is because the shape annotations are more difficult to get and thus less accurate, whereas synthetic data has perfect shape annotations. Even so, the performance gap between real and synthetic are small.

\begin{figure}
    \centering
    \includegraphics[width=\linewidth]{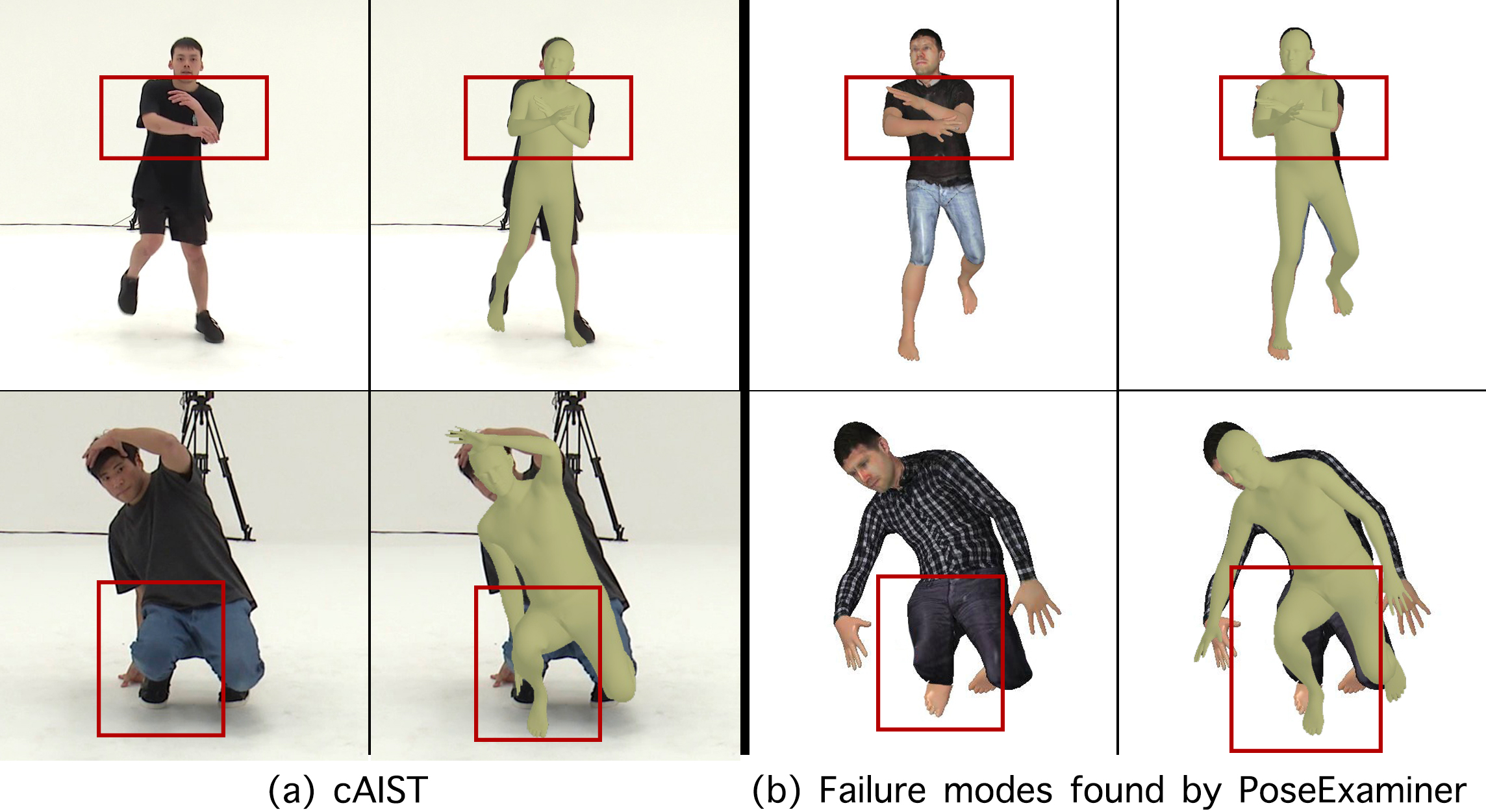}
    \caption{\textbf{Failure modes generalize well to real images.} We compare the failure modes discovered by \OURS (b) with images from the cAIST dataset (a), and visualize the results on two groups of images that have similar poses. We observe that similar poses across synthetic and real domains result in similar errors. 
    More visualizations of the failure modes are provided in Supp.}
    \vspace{-1.0em}
    \label{fig:domainGap}
\end{figure}

\noindent\textbf{Failure Modes Generalize Well to Real Images.} After running \OURS and finding failure modes of SPIN and PARE, we compare them with the real images in the cAIST-EXT dataset and find several real images with people in very similar poses. We visualize the results in Fig.~\ref{fig:domainGap}. The HPS algorithms perform similarly in these examples, and more importantly, Fig.~\ref{fig:domainGap} suggests that a method may generate the same error patterns on the same failure mode, even across the domains of synthetic and real, which \textbf{demonstrates that \OURS is able to provide insights into the model's real-world behaviors. }

\subsection{Robustness towards OOD Poses}
\label{sec:poseRobustness}
\begin{table}
\renewcommand\arraystretch{0.9} 
    \centering 
    \setlength\tabcolsep{2pt}
    \resizebox{1\columnwidth}{!}{
    \begin{tabular}{c|cccc|cc}
    \toprule
    &\multicolumn{4}{c}{Model robustness}& \multicolumn{2}{|c}{Subspace quality}\\
    \cmidrule(lr){2-5} \cmidrule(l){6-7}
    & {\small Succ. Rate $\downarrow$ }&  {\small Region Size$\downarrow$ }&  {\small meanMPJPE$\downarrow$ }&  {\small maxMPJPE$\downarrow$ }&  {\small minMPJPE}  & {\small Pnae } \\
    \midrule
    HMR~\cite{kanazawa_hmr} &  97.5\% & 8.521 & 225.03 & 448.37 & 83.55 & 0.4\% \\
    SPIN~\cite{SPIN:ICCV:2019} & 95.0\% & 5.417 & 169.64 & 352.69 & 84.14 & 0.9\% \\
    HMR-EFT~\cite{joo2021exemplar} & 84.1\% & 4.675 & 161.35 & 337.46 & 81.84 & 1.2\% \\
    PARE~\cite{kocabas2021pare} & 75.3\% & 3.948 & 150.06 & 300.76 & 81.81 & 1.3\%\\
    \bottomrule
    \end{tabular}}
    \caption{\textbf{Robustness towards adversarial examination of articulated pose.} We compare success rate, region size in radian, and MPJPE. PARE outperforms other algorithms on all these metrics, but compared to its good results on 3DPW, the high success rate and large region size indicate that it still has some weaknesses which could not be detected by standard testing.}
    \vspace{-1.0em}
    \label{table:poseStudy}
\end{table}

Articulated pose is a very important source of variability. It has a very large and high-dimensional space. Efficiently searching the articulated pose space while avoiding invalid poses is one of our main challenges. Meanwhile, despite being a very critical factor, little work has been done on the OOD poses. In this section, we investigate the robustness of four methods to articulated pose $\theta_a\in \mathbb{R}^{63}$, which also proves the efficiency and effectiveness of \OURS. 

Since we only consider articulated poses here, we use the simplest set of parameters for other factors (\ie, mean body shape, white background image, standard clothing, no occlusion, and no global rotation so that the subject is facing the camera). 
The results are shown in Tab.~\ref{table:poseStudy}

\noindent\textbf{Effectiveness of \OURS.}
The success rate in Tab.~\ref{table:poseStudy} indicates that \OURS is able to localize failure cases very efficiently. Even with the most robust method (PARE), $75.3\%$ of agents (241 agents) discover failure cases in no more than 300 iterations. Then we evaluate the boundary quality. We can see that the minMPJPE is slightly smaller than the threshold $T=90$, indicating that the agent has already found a large enough boundary for each failure case. Meanwhile, Pnae shows that there is only about $1\%$ of poses in these subspaces result in errors smaller than $T$, indicating that the boundary is accurate enough that only a few non-adversarial examples are included in the subspaces.

In short, \textbf{we show that with a total of 600 iterations, \OURS is able to efficiently search the complex high-dimensional space, discover a variety of failure modes, and accurately determine their boundaries.}
Due to space constraints, we only provide two failure modes of SPIN and PARE in Fig.~\ref{fig:domainGap} and three modes of PARE in Fig.~\ref{fig:teaser}. More examples are provided in the Supp.

\noindent\textbf{Testing HPS Methods on OOD Poses.}
From Tab.~\ref{table:poseStudy} we can find that current methods are not robust to ODD poses even when only considering articulated pose and excluding all other challenges such as occlusion, orientation, uncommon body shape, and so on. PARE is the most robust method, but it still has many failure modes and fails to give accurate estimates to some realistic but possibly less common poses.

\subsection{Robustness towards Body Shape and Occlusion}
\label{sec:otherRobustness}

As previously stated, our method can be used to study any controllable factor of the rendering pipeline. We also analyze all other factors in this work, but we only include results on body shape and occlusion in this section due to space limits. Results for global rotation, clothing, background, lighting, and gender are provided in the Supp.

\begin{table}
\renewcommand\arraystretch{0.9} 
    \centering 
    \setlength\tabcolsep{3pt}
    \resizebox{1\columnwidth}{!}{
    \begin{tabular}{c|cc|cc}
    \toprule
    &\multicolumn{2}{c}{Without occluder}& \multicolumn{2}{|c}{With small occluder}\\
    \cmidrule(lr){2-3} \cmidrule(l){4-5}
    & Succ. Rate $\downarrow$& Region Size$\downarrow$ & Succ. Rate $\downarrow$& Region Size$\downarrow$  \\
    \midrule
    HMR-EFT~\cite{joo2021exemplar} & 84.1\% & 4.675 & 100\% & 7.745  {\small (+3.070)}\\
    PARE~\cite{kocabas2021pare} & 75.3\% & 3.948 & 94.5\% & 5.522  {\small (+1.574)}\\
    \bottomrule
    \end{tabular}}
    \caption{\textbf{Robustness towards occlusion.} PARE is more robust to occlusion compared with HMR-EFT.}
    \vspace{-1.0em}
    \label{table:occStudy}
\end{table}
\begin{table*}
\renewcommand\arraystretch{0.9}
    \centering
    \setlength\tabcolsep{3pt}
    \resizebox{1\textwidth}{!}{
    \begin{tabular}{l|ll|lll|ll|ll}
    \toprule
    &\multicolumn{2}{c|}{Robustness (\OURS)}&\multicolumn{3}{c|}{ID dataset (3DPW)}&\multicolumn{2}{c|}{OOD dataset (cAIST)}&\multicolumn{2}{c}{Extreme poses (cAIST-EXT)}\\
    \cmidrule(lr){2-3} \cmidrule(lr){4-6} \cmidrule(lr){7-8}\cmidrule(l){9-10}
    & Succ. Rate$\downarrow$ & Region Size$\downarrow$  & MPJPE$\downarrow$ & PA-MPJPE$\downarrow$ & PVE$\downarrow$  & MPJPE$\downarrow$ & PA-MPJPE$\downarrow$ & MPJPE$\downarrow$ & PA-MPJPE$\downarrow$  \\
    \midrule
    SPIN~\cite{SPIN:ICCV:2019} & 95.0\% & 5.417 & 94.11 & 57.54 & 111.12 & 108.09 & 68.59 & 133.65 & 81.03 \\
    \ + PE (Ours)  & 76.6\% \de{-18.4} & 3.964 \de{-1.435}& 88.66 \de{-5.45} & 54.34 \de{-3.20} & 103.94 \de{-7.18} & 98.88 \de{-9.21} & 65.28 \de{-3.31} & 120.98 \de{-12.67} & 76.82 \de{-4.21} \\
    \midrule
    HMR-EFT~\cite{joo2021exemplar} & 84.1\% & 4.675 & 85.23 & 51.88 & 107.88 & 98.55 & 66.01 & 122.19 & 78.39\\
    \ + PE (Ours)  & 63.6\% \de{-20.5}& 3.429 \de{-1.246}& 80.41 \de{-4.82} &49.15 \de{-2.73} &101.43 \de{-6.45} &88.53 \de{-10.02} & 64.00 \de{-2.01} & 112.78 \de{-9.41} & 75.15 \de{-3.24}   \\
    \midrule
    PARE~\cite{kocabas2021pare} & 75.3\% & 3.948 & 81.81 & 50.78 & 102.27 & 99.15 & 62.43 & 117.45 & 72.68   \\
    \ + PE (Ours)  & 48.4\% \de{-26.9} & 2.953 \de{-0.995} & 77.46 \de{-4.35} & 48.01 \de{-2.77} & 94.86 \de{-7.41} & 87.44 \de{-11.71} & 59.80 \de{-2.63} & 109.82 \de{-7.63} & 70.73 \de{-1.95}\\
    \bottomrule
    \end{tabular}}
    \vspace{-0.8em}
    \caption{\textbf{Fine-tuning with \OURS.} `PE' is short for \OURS. Despite being synthetic, fine-tuning on the failure modes discovered by \OURS can significantly improve the robustness and real-world performance of all methods on all benchmarks.}
    \vspace{-0.5em}
    \label{table:training}
\end{table*}

\begin{table*}
\renewcommand\arraystretch{0.9}
    \setlength\tabcolsep{5pt}
    \centering
    \resizebox{1\textwidth}{!}{
    \begin{tabular}{l|cc|ccc|cc|cc}
    \toprule
    &\multicolumn{2}{c|}{Robustness (\OURS)}&\multicolumn{3}{c|}{ID dataset (3DPW)}&\multicolumn{2}{c|}{OOD dataset (cAIST)}&\multicolumn{2}{c}{Extreme poses (cAIST-EXT)}\\
    \cmidrule(lr){2-3} \cmidrule(lr){4-6} \cmidrule(lr){7-8}\cmidrule(l){9-10}
    & Succ. Rate$\downarrow$ & Region Size$\downarrow$  & MPJPE$\downarrow$ & PA-MPJPE$\downarrow$ & PVE$\downarrow$  & MPJPE$\downarrow$ & PA-MPJPE$\downarrow$ & MPJPE$\downarrow$ & PA-MPJPE$\downarrow$ \\
    \midrule
    PARE~\cite{kocabas2021pare} & 75.3\% & 3.948 & 81.81 & 50.78 & 102.27 & 99.15 & 62.43 & 117.45 & 72.68 \\
    PARE + PE (random) & 69.5\% & 3.542 & 79.77 & 50.20 &  97.80 & 95.10 & 62.40 & 115.54 & 73.25 \\
    PARE + PE (easy) & 65.4\% & 3.373 & \textbf{74.44} & \textbf{47.86} & \textbf{92.94} & 105.09 & 59.95 & 129.77 & 71.84  \\
    PARE + PE (hard) & \textbf{46.7\%} & \underline{3.005} & 81.15 & 48.54 & 100.08 & \textbf{86.92} & \textbf{59.03} & \textbf{107.13} & \textbf{70.15} \\
    PARE + PE (mixed)  & \underline{48.4\%} & \textbf{2.956} & \underline{77.46} & \underline{48.01} & \underline{94.86} & \underline{87.44} & \underline{59.80} & \underline{109.82} & \underline{70.73} \\
    \bottomrule
    \end{tabular}}
    \vspace{-0.8em}
    \caption{\textbf{Ablation study.} `Random' is short for `random sampling'. 
    \OURS with different difficulty levels leads to different results on different real-world benchmarks. Best in \textbf{bold}, second best \underline{underlined}.}
    \vspace{-0.5em}
    \label{table:abl}
\end{table*}

\begin{figure}
    \centering
    \includegraphics[width=\linewidth]{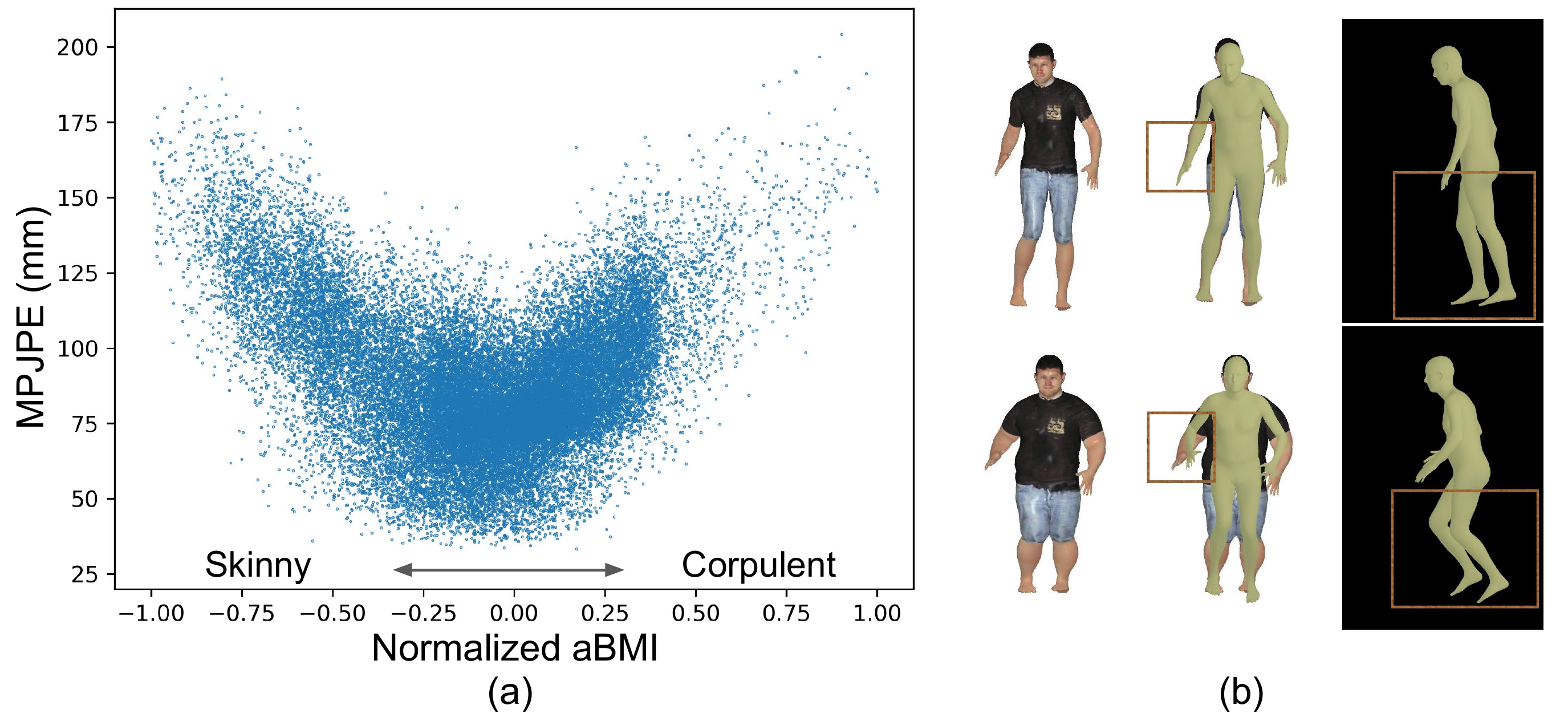}
    \caption{\textbf{Body shape bias on PARE.} (a) shows the relations between MPJPE and aBMI of all samples during optimization and (b) shows the human meshes of one agent in the first and final iterations and the estimates given by PARE.
    We observe a large drop in performance for the current SOTA method on humans with skinny and corpulent body shapes.}
    \vspace{-1.0em}
    \label{fig:shapeBias}
\end{figure}
\noindent\textbf{Shape.} To study the robustness to body shape, we consider two experiments: (1) only optimizing shape parameters $\beta\in\mathbb{R}^{10}$ while the pose $\theta_a\in \mathbb{R}^{63}$ are randomly initialized and fixed, and (2) optimizing both shape and articulated pose simultaneously. We initialize and freeze other parameters following Sec.~\ref{sec:poseRobustness}. We find that all agents eventually converge on either a skinny or a corpulent human body. To quantify the results, we borrow the concept of Body Mass Index (BMI) that measures leanness or corpulence based on height and weight, and compute approximate Body Mass Index (aBMI): $aBMI=volume/height^2$. Given a shape parameter, we render a human mesh with the canonical pose, and then we can read the height and use~\cite{zhang2001efficient} to compute its volume and get the aBMI of this shape. Fig.~\ref{fig:shapeBias} shows the relations between MPJPE and aBMI of all samples during optimization, as well as some visualization results.

\noindent\textbf{Occlusion.}
\label{sec:occRobustness}
Occlusion is a popular problem in HPS~\cite{kocabas2021pare,3doh}. We optimize the pose parameters $\theta_a\in \mathbb{R}^{63}$ and occlusion parameters $t_{occ}\in\mathbb{R}^3$ simultaneously and compared the robustness of PARE and HMR-EFT. Results are shown in Tab.~\ref{table:occStudy}. PARE outperforms HMR-EFT in robustness, but in some hard poses, even minor occlusion may cause large errors. (See Supp. for visualizations.) This implies that current methods solve occlusion partly based on the pose distribution of training data. The problem becomes worse when a person with an OOD pose is occluded.

\subsection{Improving HPS Methods with \OURS}
\label{sec:train}
To evaluate the efficacy of fine-tuning with \OURS, we fine-tune the pretrained models of SPIN, HMR-EFT, and PARE in Tab.~\ref{table:training}. 
Training with the failure modes discovered by \OURS is very effective. It significantly improves robustness and performance on both ID and OOD datasets. Note that the failure modes we learned here mainly focus on hard and extreme poses, we are surprised to see that performance on the 3DPW dataset is improved as well. In fact, we can use a simpler examiner to further boost the performance on 3DPW (See Sec.~\ref{sec:ablStudy}).

\subsection{Ablation Study}
\label{sec:ablStudy}
\noindent\textbf{Training with Different Settings of \OURS.}
As mentioned above, fine-tuning PARE with \OURS of different difficulty levels yields different results. We define two levels of difficulty here: easy and hard. We set a more strict restriction to the easy examiner so that it can only search a very small parameter space, whereas the hard examiner searches a much larger parameter space. We also use curriculum learning to combine these two examiners and name it `mixed’ (\ie the \OURS used in Tab.~\ref{table:training}). The results are provided in Tab.~\ref{table:abl}. Fine-tuning PARE with easy \OURS significantly improves performance on 3DPW but degrades performance on cAIST, while hard \OURS benefits more on the cAIST dataset. This suggests that a model with larger capability will benefit more from fine-tuning with \OURS.

\noindent\textbf{Comparison to Random Sampling.} 
We compare our RL-based searching with random sampling in Tab.~\ref{table:abl}. For random sampling, instead of sampling parameters from the discovered failure modes, we directly sample the same number of parameters from the entire space. Random sampling also improves performance on all benchmarks. However, \OURS (`mixed') improves the performance and robustness by a much larger margin, especially on OOD datasets, demonstrating the efficacy of training on failure modes.

\vspace{-0.3em}\section{Conclusion}\vspace{-0.3em}

We have introduced \OURS, a learning-based testing method that automatically diagnoses HPS algorithms by searching over the parameter space of human pose images for failure modes. These failure modes are relevant in real-world scenarios but are missed by current benchmarks.
\OURS enables a highly efficient way to study and quantify the robustness of HPS methods towards various factors such as articulated pose, thereby providing new insights for existing methods.
Furthermore, despite being synthetic, fine-tuning on the failure modes discovered by \OURS can significantly improve the robustness and real-world performance on all benchmarks.

\textbf{Acknowledgements.} AK acknowledges support via his Emmy Noether Research Group funded by the German Science Foundation (DFG) under Grant No. 468670075. QL acknowledges support from the NIH R01 EY029700 and Army Research Laboratory award W911NF2320008. This research is also based upon work supported in part by the Office of the Director of National Intelligence (ODNI), Intelligence Advanced Research Projects Activity (IARPA), via [2022-21102100005]. The views and conclusions contained herein are those of the authors and should not be interpreted as necessarily representing the official policies, either expressed or implied, of ODNI, IARPA, or the U.S. Government. The US. Government is authorized to reproduce and distribute reprints for governmental purposes notwithstanding any copyright annotation therein.

\newpage
{\small
\bibliographystyle{ieee_fullname}
\bibliography{egbib}
}
\newpage
\appendix
\section*{Appendices}
Here we provide details and extended experimental results omitted from the main paper for brevity. Sec.~\ref{sec:Hyper} gives hyperparameters of \OURS used for testing and fine-tuning HPS algorithms. Sec.~\ref{sec:extexp} contains extended experimental evaluations. Sec.~\ref{sec:vis} gives more visualization results. Sec.~\ref{sec:sync3DPW} and Sec.~\ref{sec:aist} provide details to generate the synthetic 3DPW dataset and to filter out wrong annotations from the AIST++ dataset. Sec.~\ref{sec:pseudocode} provides pseudocodes of \OURS, and Sec.~\ref{sec:lim} discusses limitations and future work.
\section{Hyperparameters of \OURS}
\label{sec:Hyper}
In our main paper, we use three sets of hyperparameters that are corresponding to three different difficulty levels.

\noindent\textbf{Standard.} We use human clothing ID 1 (see Fig.~\ref{fig:text}) and plain white background. We set adversarial threshold $T=90$, and the limits of global rotation along the Y/X/Z-axis are $\pm 0.02\pi, \pm 0.02\pi, \pm 0.02\pi$, respectively. The searching boundary of policy $\pi_\omega$ is $[-2,2]$. The standard difficulty level is used to evaluate the robustness of current methods towards articulated poses, shape, lighting, and occlusion. For other evaluation experiments such as clothing, background, and global rotation, we change the corresponding factor (\eg the human texture ID, or the limits of global rotation) and keep the rest.

\noindent\textbf{Easy.} We use the 5 most distinguishable human clothing and the 10 most distinguishable backgrounds (see the project page). We set adversarial threshold $T=80$ and the searching boundary of policy $\pi_\omega$ is $[-1.5,1.5]$. The limits of global rotation along the Y/X/Z-axis are $0, \pm 0.05\pi, 0$, respectively. \OURS with the easy difficulty level generates good performance when it is used to fine-tune the model tested on the 3DPW dataset.

\noindent\textbf{Hard.} We use 12 hardly distinguishable human clothing and the 5 most difficult backgrounds (see the project page). We set adversarial threshold $T=90$ and the searching boundary of policy $\pi_\omega$ is $[-3,3]$. The limits of global rotation along the Y/X/Z-axis are $\pm 0.4\pi, \pm 0.05\pi, \pm 0.05\pi$, respectively. \OURS with the hard difficulty level generates good performance when it is used to fine-tune the model tested on the cAIST and cAIST-EXT dataset.

\noindent\textbf{Mixed.} To achieve good performance on both IID and OOD datasets, we mix these three difficulty levels during fine-tuning. Specifically, in the first two epochs, we use the easy \OURS, and in the following two epochs, we use the standard one to search for failure modes, and in the rest epochs, we use the hard one.

\section{Extended Experiments}
\label{sec:extexp}
\subsection{Performance Gap between Real and Synthetic cAIST-EXT datasets}

\begin{table}
\renewcommand\arraystretch{0.9}
    \centering
    \setlength\tabcolsep{3pt}
    \resizebox{0.95\columnwidth}{!}{
        \begin{tabular}{l|cc|cc}
        \toprule
        &\multicolumn{2}{c}{real cAIST-EXT}& \multicolumn{2}{|c}{sync cAIST-EXT}\\
        \cmidrule(lr){2-3} \cmidrule(l){4-5}
        & MPJPE$\downarrow$ & PA-MPJPE$\downarrow$ & MPJPE$\downarrow$ & PA-MPJPE$\downarrow$  \\
        \midrule
        SPIN~\cite{SPIN:ICCV:2019} & 133.7 & 81.0 & 124.1 {\small (-9.6)} & 76.7 {\small (-4.3)} \\
        PARE~\cite{kocabas2021pare} & 117.5 & 72.7 & 111.4 {\small (-6.1)} & 69.2 {\small(-3.5)}\\
        \bottomrule
        \end{tabular}}
    \caption{\textbf{Measuring the performance gap of SPIN and PARE on real and synthetic cAIST-EXT datasets.} Real cAIST-EXT only has pseudo labels while the synthetic version directly uses ground-truth human meshes. Therefore, the annotations of sync cAIST-EXT are very accurate, but on the real cAISt-EXT, the annotations are less accurate. This difference increases the performance gap. Nonetheless, the performance gap between real and synthetic are not large and even favorable towards the synthetic data.}
    \label{table:realSyn-aist}
\end{table}
We provide the results of SPIN and PARE on real and synthetic cAIST-EXT datasets in Tab.~\ref{table:realSyn-aist}. Both methods achieve similar performance on the real and synthetic versions. They even achieve better performance on the synthetic one. One reason for this is that the real cAIST-EXT only has pseudo labels that are less accurate, while the synthetic cAIST-EXT has accurate labels since the images are directly rendered using the ground-truth human meshes. In short, the performance gap is small enough that the synthetic data can be used to evaluate the performance of an HPS method trained on real images.

\subsection{Ablation on Number of Samples}
\begin{table}
    \setlength\tabcolsep{3pt}
    \small 
    \centering
    \resizebox{\columnwidth}{!}{
    \begin{tabular}{lccccc}
    \toprule
     & Pnae & minMPJPE & maxMPJPE & meanMPJPE & medianMPJPE \\
    \midrule
    50 samples & 0.8\% & 199.69 & 213.52 & 151.97 & 149.31 \\
    200 samples & 1.30\% & 81.81 & 300.76 & 150.06 & 148.23 \\
    5000 samples & 1.32\% & 79.43 & 305.52 & 150.41 & 148.30 \\
    \bottomrule
    \end{tabular}}
    \caption{\textbf{Ablation on number of samples.} 200 samples give very similar results compared with 5000 samples.}
    \label{rebut:numofsample}
\end{table}

Note that the HPS model is usually an end-to-end deep network of which the performance is unexplainable, and the pose space is extremely high-dimensional. Therefore, in our paper, to better understand and evaluate the failure modes, we use uniform sampling to learn the property of each subspace. We made an additional ablation in Tab.~\ref{rebut:numofsample} regarding the number of required samples, and observe that 200 samples per subspace already provide a good estimate of the properties of a failure mode.

\subsection{Ablation on Joint Ranges}
\begin{table}
    \setlength\tabcolsep{5pt}
    \small 
    \centering
    \resizebox{\columnwidth}{!}{
    \begin{tabular}{lcccc}
    \toprule
     & 20$\%$ AMASS & 50$\%$ AMASS & 100$\%$ AMASS & Physical limits  \\
     \midrule
     Region Size & 1.738 & 2.649 & 3.212 & 3.948 \\
     meanMPJPE & 114.05 & 130.67 & 139.21 & 150.06 \\
    \bottomrule
    \end{tabular}}
    \caption{\textbf{Robustness of PARE under different joint ranges.} We report region size and meanMPJPE here.}
    \label{rebut:jointrange}
\end{table}

So far, when studying the robustness to OOD poses, we only consider the physical limits. It will generate many uncommon poses.
One way to address this issue is to set different joint ranges for searching, from a very small region that only includes very common poses to bigger ranges that also contain unusual poses. 
We made an additional experiment (Tab.~\ref{rebut:jointrange}) where we estimate the pose distribution in the AMASS dataset, and set joint ranges that cover $20\%$, $50\%$, and $100\%$ poses respectively. 
However, we want to emphasize the importance of testing uncommon poses, as algorithms must be robust to such edge cases in practice.

\subsection{Robustness towards Clothing, Lighting, Background and Global Rotation}

\begin{table}
\renewcommand\arraystretch{0.9}
    \small 
    \centering
    \setlength\tabcolsep{2pt}
    \resizebox{1\columnwidth}{!}{
    \begin{tabular}{l|ccc|ccc}
    \toprule
    &\multicolumn{3}{c}{sync 3DPW}& \multicolumn{3}{|c}{\OURS}\\
    \cmidrule(lr){2-4} \cmidrule(l){5-7}
    & MPJPE$\downarrow$ & PA-MPJPE$\downarrow$ & PVE$\downarrow$ & Succ. Rate$\downarrow$ & Region Size$\downarrow$ & meanMPJPE$\downarrow$  \\
    \midrule
    ID 1 & 78.3 & 52.0 & 93.2 & 74.2\% & 3.822 & 150.76 \\
    ID 17 & 80.0 & 54.5 & 92.8 & 82.0\% & 4.540 & 155.94 \\
    ID 22 & 81.4 & 53.3 & 95.0 & 80.5\% & 4.010 & 153.58 \\
    \red{ID 25} & \red{84.3} & \red{55.0} & \red{98.9} & \red{88.2\%} & \red{4.632} & \red{160.16} \\
    ID 31 & 80.9 & 52.7 & 94.5 & 82.8\% & 4.681 & 156.49 \\
    \bottomrule
    \end{tabular}}
    \caption{\textbf{Robustness towards clothing.} Here we show results on the five most representative clothing (Fig.~\ref{fig:text}). Although human clothing makes little difference in simple poses, the robustness of PARE still decreases when less distinguishable clothes (\eg \red{ID25}) are used. However, in general, compared to other factors, PARE is relatively robust towards common clothing with no adversarial noises.}
    \label{table:textureStudy}
\end{table}
\begin{figure}
    \centering
    \includegraphics[width=\linewidth]{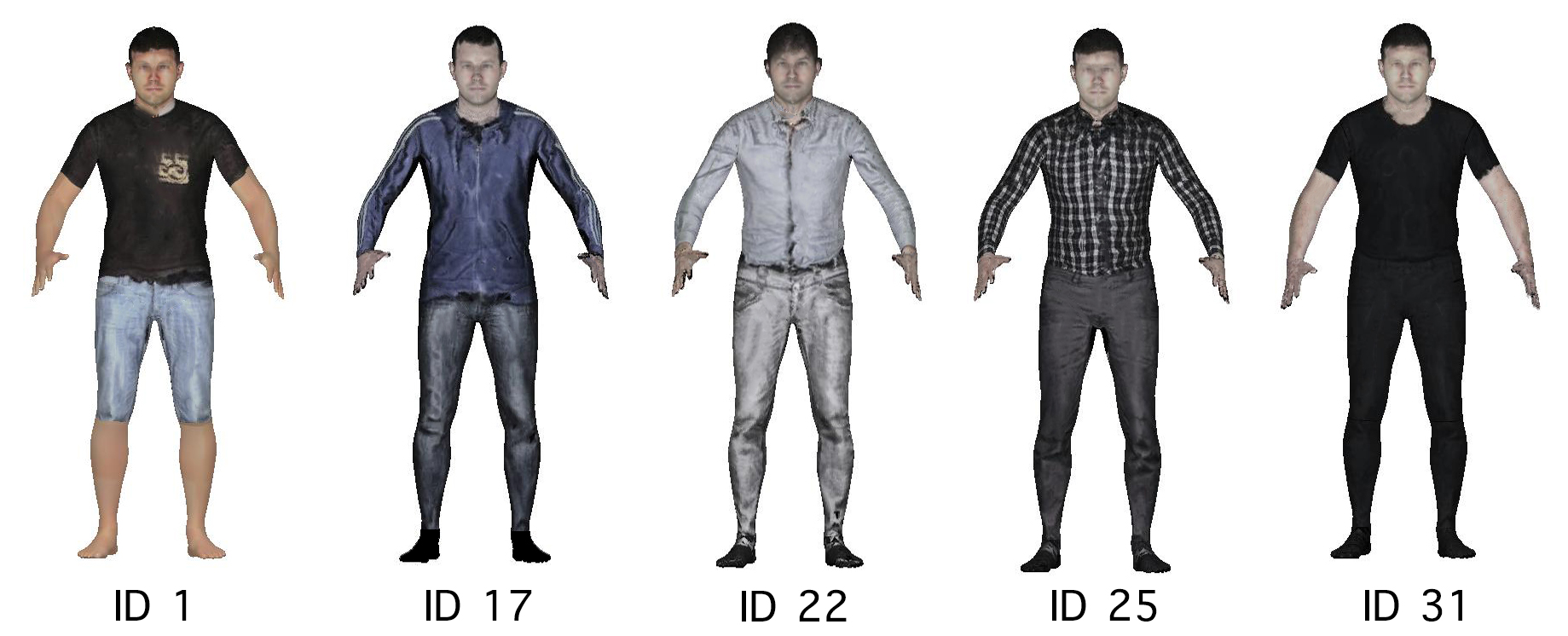}
    \caption{\textbf{Visualizations of 5 human clothing.} }
    \label{fig:text}
\end{figure}
\noindent\textbf{Clothing.}
To study the robustness of PARE to human clothing, we first design a preliminary experiment. 40 high-quality UV maps are selected to generate human meshes with different clothes. Then we directly generate 40 synthetic 3DPW datasets, one for each UV map. The left half of Tab.~\ref{table:textureStudy} shows the results on the 5 most representative textures (Fig.~\ref{fig:text}). If we only consider common clothing with no (adversarial) noises added to them, the textures we selected do not make a large difference.

Then we study the robustness of current methods to human clothing in extreme/hard poses. Since we do not directly optimize the parameters that are used to generate UV maps, we use our \OURS to learn weaknesses regarding articulated poses but with different UV maps. The results are provided in the right half of Tab.~\ref{table:textureStudy}. Although clothing makes little difference in simple poses, certain poses can be difficult in some clothes but simple in others. With less distinguishable clothes, the robustness of current methods will decrease. However, in general, compared to other factors, PARE is robust towards common clothing with no adversarial noises.

\begin{figure}
    \centering
    \includegraphics[width=\linewidth]{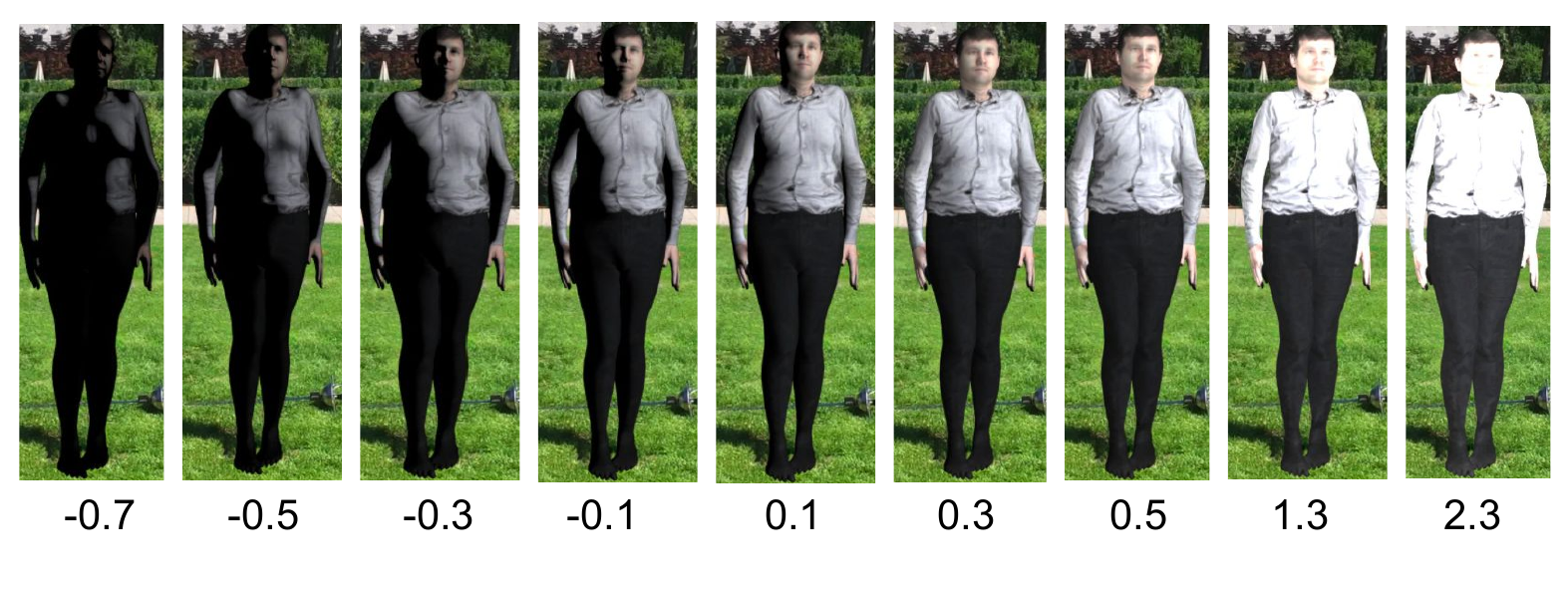}
    \caption{\textbf{Visualizations of different lighting intensities.} }
    \label{fig:lighting}
\end{figure}

\noindent\textbf{Lighting.}
Same to clothing, we study the robustness of current methods towards lighting in two manners: on synthetic 3DPW and on \OURS. We generate images with different lighting intensities (Fig.~\ref{fig:lighting}) and test PARE on them. The results are provided in Tab~\ref{table:lightingStudy}. PARE is relatively robust to overexposure but less so to underexposure.

\begin{table}
\renewcommand\arraystretch{0.9}
    \centering
    \setlength\tabcolsep{2pt}
    \resizebox{1\columnwidth}{!}{
    \begin{tabular}{c|ccc|ccc}
    \toprule
    &\multicolumn{3}{c}{sync 3DPW}& \multicolumn{3}{|c}{\OURS}\\
    \cmidrule(lr){2-4} \cmidrule(l){5-7}
    & MPJPE$\downarrow$ & PA-MPJPE$\downarrow$ & PVE$\downarrow$ & Succ. Rate$\downarrow$ & Region Size$\downarrow$ & meanMPJPE$\downarrow$  \\
    \midrule
    \red{$-0.7$} & \red{102.6} & \red{75.3} & \red{129.3} & \red{93.4\%} & \red{6.254} & \red{184.26} \\
    $-0.5$ & 88.5 & 62.8 & 110.9 & 91.8\% & 5.366 & 170.15 \\
    $-0.3$ & 87.2 & 55.5 & 107.2 & 88.3\% & 4.423 & 168.37 \\
    $-0.1$ & 85.3 & 53.0 & 104.5 & 88.9\% & 4.675 & 171.75 \\
    $0.1$ & 82.3 & 51.8 & 98.6 & 83.7\% & 4.369 & 155.75 \\
    \underline{$0.3$} & \underline{79.6} & \underline{51.7} & \underline{94.5} & \underline{78.9\%} & \underline{4.045} &  \underline{156.11} \\
    $0.5$ & 79.4 & 52.1 & 94.3 & 80.1\% & 4.247 & 160.42 \\
    $1.3$ & 79.5 & 53.3 & 93.5 & 85.4\% & 4.403 & 158.45 \\
    \blue{$2.3$} & \blue{78.9} & \blue{53.2} & \blue{93.8} & \blue{87.4\%} & \blue{4.361} & \blue{165.22}\\
    \bottomrule
    \end{tabular}}
    \caption{\textbf{Robustness towards lighting.} We generate images with different lighting intensities (Fig.~\ref{fig:lighting}). Compared with normal exposure (\underline{underline}), PARE is relatively robust to overexposure (\blue{blue}) but less so to underexposure (\red{red}). }
    \label{table:lightingStudy}
\end{table}

\begin{figure}
    \centering
    \includegraphics[width=\linewidth]{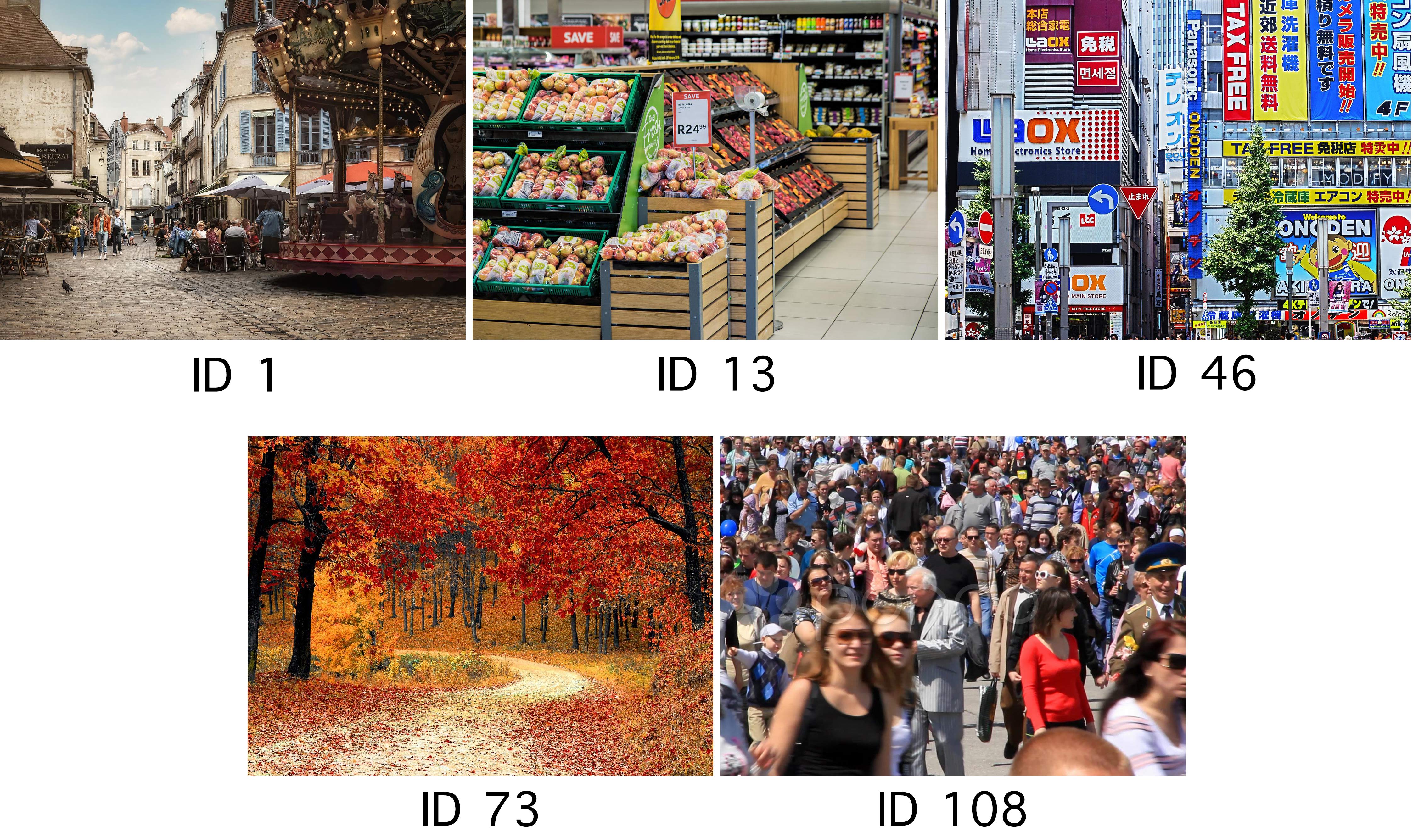}
    \caption{\textbf{Visualizations of 5 background images.} }
    \label{fig:bkgd}
\end{figure}

\noindent\textbf{Background.}
Same to clothing, we study the robustness of current methods towards different backgrounds using synthetic 3DPW and on \OURS. We show the results on five representative background images (Fig.~\ref{fig:bkgd}) in Tab.~\ref{table:bkgdStudy}. Current HPS methods such as PARE are still sensitive to crowded scenes (ID 108), even when ground-truth bounding boxes are provided and the input images are tightly cropped around the person. However, in general, PARE is robust and performs well on other common background images with no adversarial noises (ID 1, 13, 46, and 73).

\begin{table}
\renewcommand\arraystretch{0.9}
    \centering
    \setlength\tabcolsep{2pt}
    \resizebox{1\columnwidth}{!}{
    \begin{tabular}{l|ccc|ccc}
    \toprule
    &\multicolumn{3}{c}{sync 3DPW}& \multicolumn{3}{|c}{\OURS}\\
    \cmidrule(lr){2-4} \cmidrule(l){5-7}
    & MPJPE$\downarrow$ & PA-MPJPE$\downarrow$ & PVE$\downarrow$ & Succ. Rate$\downarrow$ & Region Size$\downarrow$ & meanMPJPE$\downarrow$  \\
    \midrule
    ID 1 & 81.87 & 55.41 & 98.41 & 88.3\% & 4.816 & 168.48 \\
    ID 13 & 84.43 & 58.63 & 102.98 & 89.8\% & 4.817 & 172.45 \\
    ID 46 & 80.52 & 52.67 & 94.52 & 87.5\% & 4.456 & 156.21 \\
    ID 73 & 80.51 & 51.61 & 94.24 & 80.4\% & 4.314 & 153.95\\
    \red{ID 108} & \red{119.94} & \red{77.65} & \red{148.84} & \red{96.1\%} & \red{6.900} & \red{227.57} \\
    \bottomrule
    \end{tabular}}
    \caption{\textbf{Robustness towards background.} We show results on five background images here (Fig.~\ref{fig:bkgd}). PARE is robust and performs well on common background images with no adversarial noises (ID 1, 13, 46, and 73). But it is still sensitive to crowded scenes (\red{ID 108}) even the input images are tightly cropped around the person.}
    \label{table:bkgdStudy}
\end{table}

\noindent\textbf{Global Rotation.}
We use \OURS to study the robustness by optimizing the global rotation and articulated pose simultaneously. However, we found that there exists a global maximum for each direction of global rotation. For example, for the rotation along the Y-axis, the larger the rotation angle, the more serious self-occlusion can happen. Therefore, to avoid the algorithm converging to poses in which the subject is nearly back to the camera and standing upside down. We set different limits of joint angles for experiments, and study only one direction every time. The results are provided in Tab.~\ref{table:gloRotStudy}. Human orientation can cause self-occlusion, leading to large errors. PARE is also sensitive to camera angles (up-down angle and tilt angle).

\begin{table}
\renewcommand\arraystretch{0.9}
    \centering
    \setlength\tabcolsep{5pt}
    \resizebox{\columnwidth}{!}{
    \begin{tabular}{ccccc}
    \toprule
    & & Succ. Rate$\downarrow$ & Region Size$\downarrow$ & meanMPJPE$\downarrow$  \\
    \midrule
     & 0 & 74.2\% & 3.822 & 150.76\\
    \midrule
    \multirow{4}{*}{\shortstack{Y-axis}}&$\pm\  1/4\ \pi$ & 75.5\% & 3.687 & 144.25\\
    &$\pm\  1/2\ \pi$ & 80.3\% & 4.024 & 158.34\\
    &$\pm\  3/4\ \pi$ & 84.1\% & 4.630 & 166.46\\
    &$\pm\  \pi$ & 90.4\% & 5.041 & 175.82 \\
    \midrule
    \multirow{2}{*}{\shortstack{X-axis}}&$\pm\  1/4\ \pi$ & 93.2\% & 5.868 & 177.56\\
    &$\pm\  1/2\ \pi$ & 96.2\% & 6.904 & 214.61\\
    \midrule
    \multirow{2}{*}{\shortstack{Z-axis}}&$\pm\  1/4\ \pi$ & 88.2\% & 4.756 & 170.64\\
    &$\pm\  1/2\ \pi$ & 92.7\% & 5.131 & 175.38 \\
    \bottomrule
    \end{tabular}}
    \caption{\textbf{Robustness towards global rotation.} Human orientation (Y-axis) can cause self-occlusion, leading to large errors. PARE is also sensitive to camera angles, including up-down angle (X-axis) and tilt angle (Z-axis).}
    \label{table:gloRotStudy}
\end{table}

\section{Visualization Results}
\label{sec:vis}
\begin{figure}
    \centering
    \includegraphics[width=\linewidth]{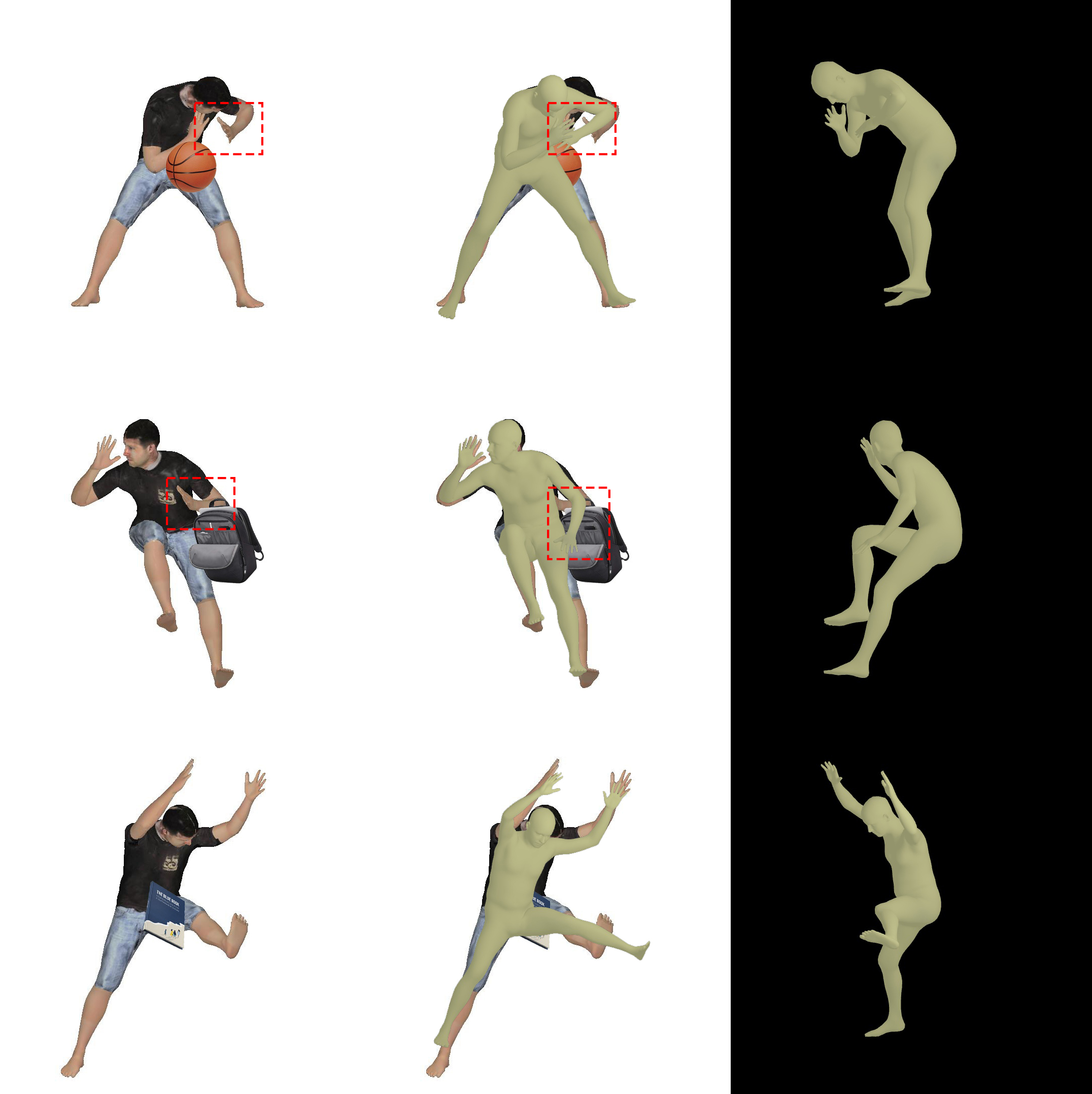}
    \caption{\textbf{Robustness towards occlusion.} PARE is designed to handle occlusion and it performs well on simple poses with occlusion. However, when considering more complex poses, a small occluder may still cause large errors.}
    \label{fig:supp_occ}
\end{figure}

\noindent\textbf{Occlusion.} We visualize several failure modes of PARE caused by occlusion in Fig.~\ref{fig:supp_occ}. As we mentioned in the main paper, PARE is robust towards occlusion in simple and IID poses. However, in some hard poses, even minor occlusion can cause large errors.

\begin{figure*}
    \centering
    \includegraphics[width=\linewidth]{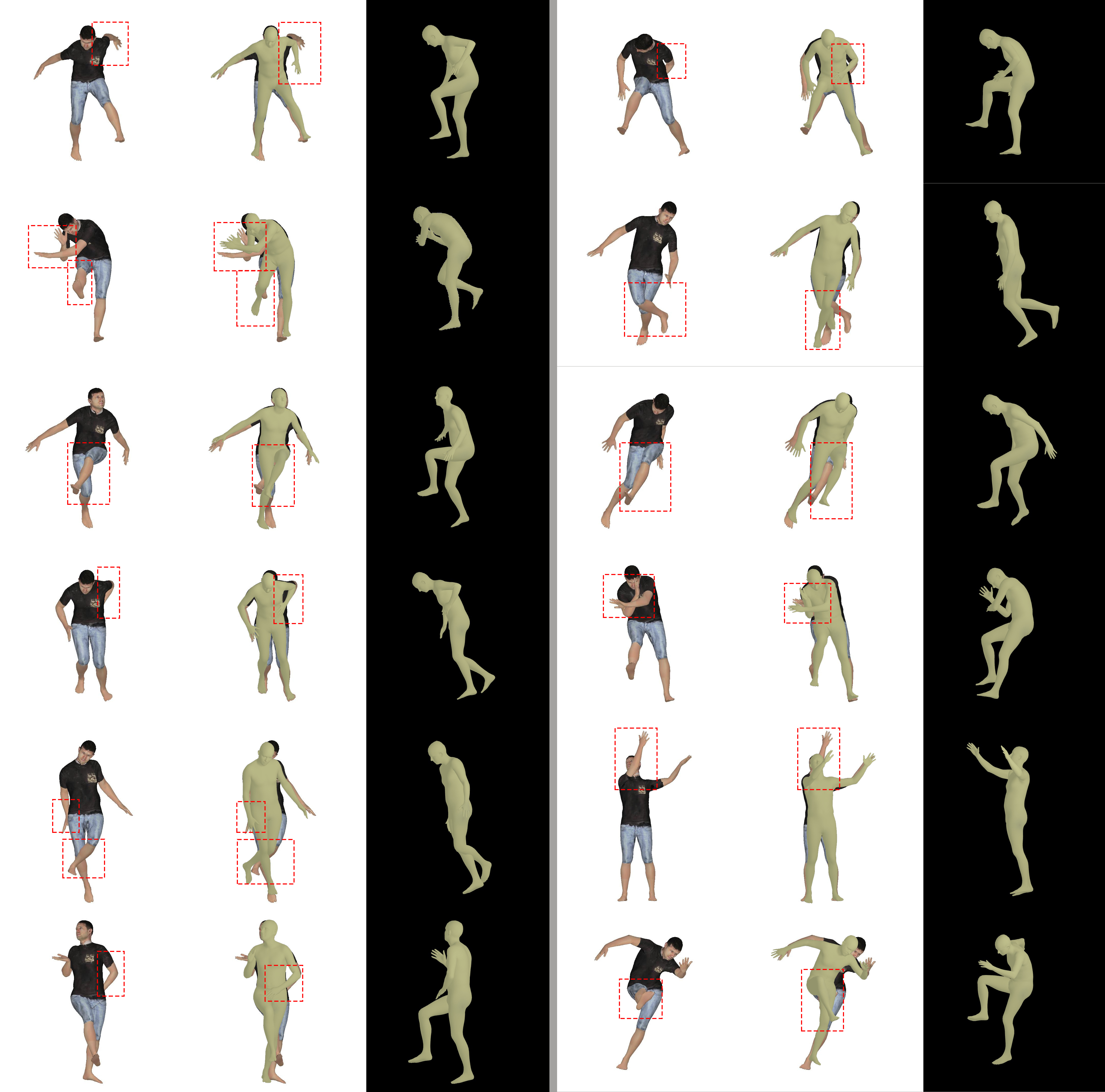}
    \caption{\textbf{Failure modes of PARE discovered by \OURS.} Our examiner finds a variety of failure modes on articulated pose that are realistic and cause large 2D errors.}
    \label{fig:supp_failure}
\end{figure*}

\noindent\textbf{Failure Modes of PARE.}
We visualize several failure modes of PARE discovered by \OURS in Fig.~\ref{fig:supp_failure}. \OURS is able to find a variety of failure modes that are realistic and cause large 2D and 3D errors.

\section{Generating Synthetic 3DPW Dataset}
\label{sec:sync3DPW}
\begin{figure*}
    \centering
    \includegraphics[width=\linewidth]{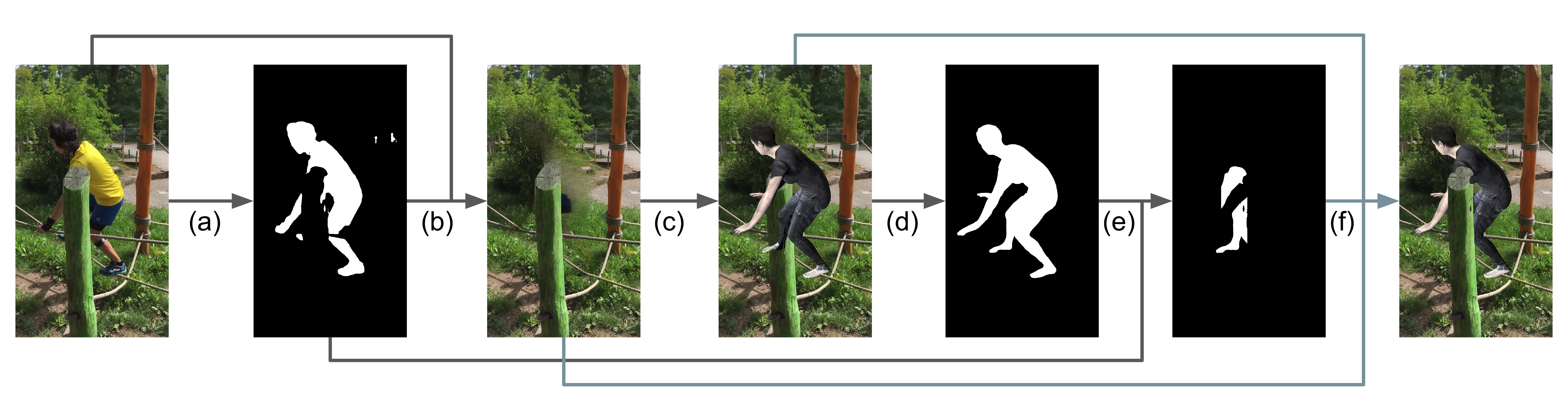}
\caption{\textbf{Generating synthetic 3DPW dataset.} We give a detailed introduction to each step in Sec.~\ref{sec:sync3DPW}.}
    \label{fig:gen_syn3dpw}
\end{figure*}
Fig.~\ref{fig:gen_syn3dpw} provides a step-by-step illustration to generate the synthetic 3DPW dataset. It has six steps:

\begin{enumerate}[(a)]
\item Use the state-of-the-art video instance segmentation method (IDOL~\cite{wu2022defense}) to generate the mask of humans.
\item Use the image imprinting method (LAMA~\cite{suvorov2022resolution}) to fill in the gaps left by the removal of the person to get the background image.
\item Use ground-truth labels to render human mesh onto the background image.
\item Use instance segmentation method (IDOL) to get the mask of synthetic human.
\item Compute the overlap of the masks of real human and synthetic human
\item Use the texture of the object on the synthetic image to restore the occlusions.
\end{enumerate}

We also follow the same steps (step a - c) to generate synthetic cAIST-EXT. Note that step d, e, and f are omitted since images in cAIST-EXT do not include occlusions.

\section{Filtering out Wrong Annotations from the AIST++ Dataset}
\label{sec:aist}
\begin{figure}
    \centering
    \includegraphics[width=\linewidth]{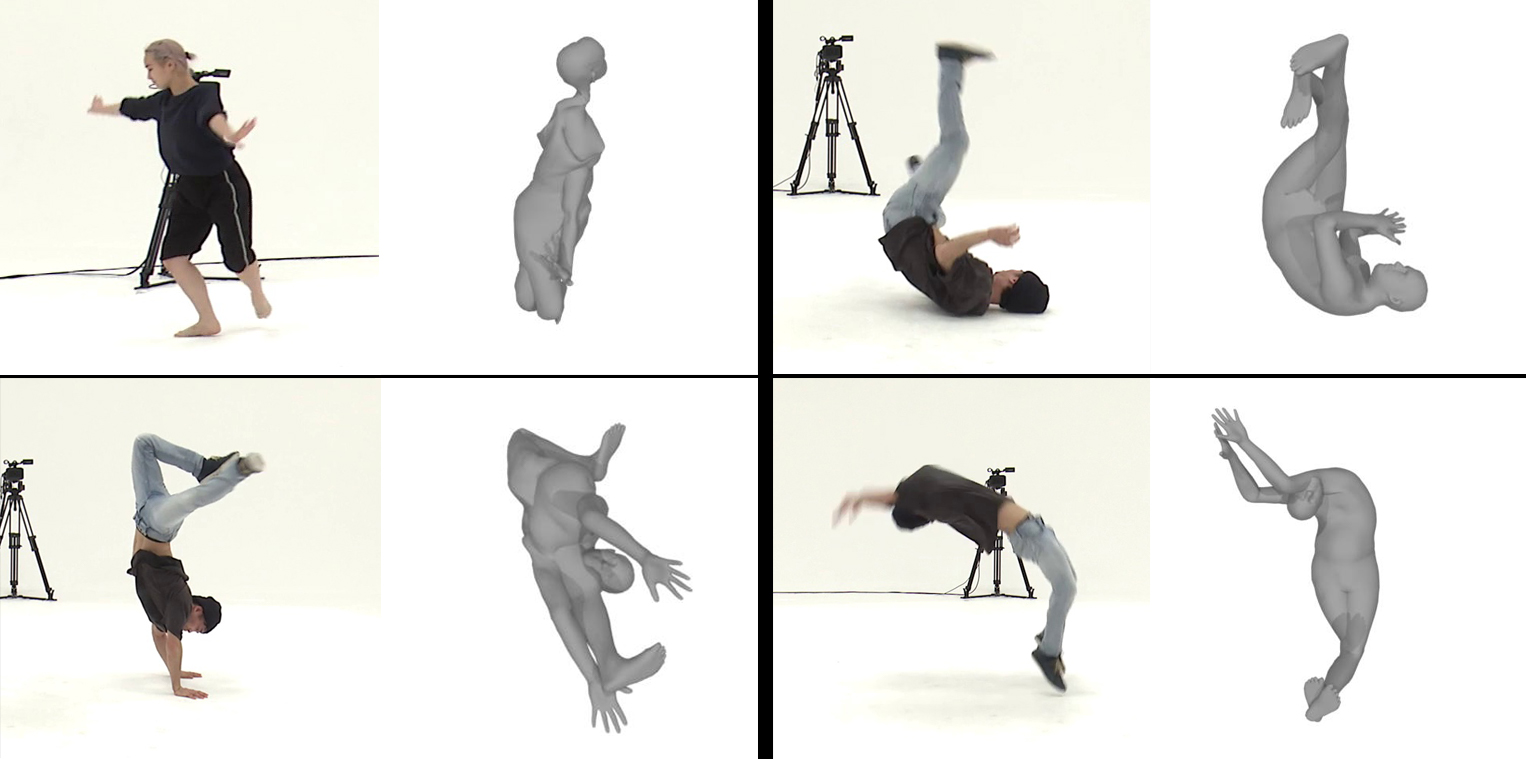}
    \caption{\textbf{Incorrect Annotations in AIST++ dataset.} Each group of images contains the original image on the left and the corresponding annotation on the right.}
    \label{fig:aist_wrong}
\end{figure}
The AIST++~\cite{aist,aist++} dataset contains pseudo labels generated from nine cameras surrounding the subjects. The annotations are relative accurate for simple poses. However, for some extreme or hard poses, the annotations are incorrect (Fig.~\ref{fig:aist_wrong}). We filter out the images with incorrect annotations in three step:
\begin{enumerate}[(1)]
\item We find that the 2D keypoint annotations are much more accurate than the 3D keypoint annotations, and the latter is more accurate than the SMPL annotations. therefore, we use the SMPL annotations to regress 3D keypoints, and then project them into 2D to get 2D keypoints. Then we check the consistency between the provided 2D annotations and the regressed 2D keypoints, as well as the provided 3D keypoint annotations and the regressed 3D keypoints.

\item We find that the main error comes from 3D estimation, or more specifically, depth estimation. For most images with correct 3D annotations, the SMPL annotations are broadly right. However, there exists some 3D skeletons that have wrong annotations on the depth of some joints. These wrong 3D skeletons, or more specifically, the abrupt changes along the depth direction, give incorrect SMPL parameters. Therefore, we check the smoothness of depth estimations and filter out the images with annotations that contain discontinuous depth estimates or impossible depth ranges.

\item The third constraint we use is the 3D joints on the face. Unlike other parts of the human body, such as arms that have a large range of motion, joints on the face, such as the eyes and nose, usually have a relatively fixed position. However, their estimates are also error-prone due to the self-occlusion. Therefore, we check the position of joints on the face and filter out the images with annotations that have impossible distances between face joints.

\end{enumerate}
After that, we randomly select 1K images and check the SMPL labels to ensure that the remaining images have correct annotations. Then we get a clean AIST++ dataset and name it cAIST.

\section{Pseudocode}
\label{sec:pseudocode}
We provide the pseudocode of phase 1 (Algo.~\ref{alg1}) and phase 2 (Algo.~\ref{alg2}) of \OURS, and the pseudocode of fine-tuning HPS methods with \OURS (Algo.~\ref{alg3}) as follows.
\begin{algorithm} 
\caption{: Finding the worst-case poses} 
\label{alg1} 
\begin{algorithmic}
\REQUIRE $\pi_{\omega^i}$: policy, $H$:a given HPS model, $S$: simulator, $f$:VPoser decoder, $\Psi^i$: other controllable parameters.
\STATE Initialize baseline $b=0.5$
\FOR{$t = 1,2,...$}
\STATE Sample $K$ latent parameters $z^i\sim \pi_{\omega^i}(z^i)$
\STATE Generate $K$ pose parameters $\theta^i_a=f(z^i)$
\STATE Render $K$ images $I^i=S(\theta^i_a, \Psi^i)$
\STATE Test $H$ on $I^i$ and obtain mean error $err^i_{2D,t}$,$err^i_{3D,t}$
\IF{$\frac{1}{10}\sum^{t}_{j=t-9}err^i_{3D,t}>T$}
\STATE Terminate and output $\omega^i$
\ENDIF
\STATE Compute rewards $R(z^i)\leftarrow c-err^i_{2D}$
\STATE Compute mean population distance $D(\pi_{\omega^i},\pi_{\omega^b})$
\STATE Update $\omega^i$ by gradient descent for maximizing\\ $\ \ \  L(\omega^i)=\mathbb{E}_{z^i\sim \pi_{\omega^i}}[R(z^i)] + \mathds{1}_{\{i\neq b\}}\gamma\mathbb{E}[D(\pi_{\omega^i},\pi_{\omega^b})]$
\STATE Update baseline $b\leftarrow(1-\tau)b+\tau R(z^i)$
\ENDFOR
\end{algorithmic} 
\end{algorithm}

\begin{algorithm} 
\caption{: Determining boundaries of the failure modes}
\label{alg2} 
\begin{algorithmic}
\REQUIRE $\theta^i_a$: adversarial point, $H$:a given HPS model, $S$: simulator, $\Psi^i$: other controllable parameters.
\STATE Initialize $\phi^i_{up}=\phi^i_{low}=\vec{0}$, $\delta=0.05$
\FOR{$t = 1,2,...$}
\STATE Select joint $j$ and sample $m$ poses $\theta^i_{a,j}\sim\mathcal{U}(\cdot)$
\STATE Render $m$ images $I^i=S(\theta^i_{a,j}, \Psi^i)$
\STATE Test $H$ on $I^i$ and obtain minimum error $err^i_{3D}$
\IF{$err^i_{3D}>T$}
\STATE Check pose possibility and update the boundary of 
\STATE \quad rotation directions of joint $j$ that yield valid pose: \\$\quad \phi^i_{up,j}\leftarrow\phi^i_{up,j}+\delta$ \ or \ $\phi^i_{low,j}\leftarrow\phi^i_{low,j}+\delta$
\STATE Update $\delta\leftarrow min\{0.001\times(err^i-T)+0.005,0.05\}$
\ENDIF
\ENDFOR
\end{algorithmic} 
\end{algorithm}

\begin{algorithm} 
\caption{ : Fine-tuning with \OURS} 
\label{alg3} 
\begin{algorithmic}
\REQUIRE $E$: pose examiner, $H$: a given HPS model, $L$: ordered list of hyperparameters, $\mathcal{T}$: original training set.
\STATE Initialize $\mathcal{F}\leftarrow \varnothing$
\FOR{$loop = 1,2,...$}
\STATE Initialize $E$ with a group for hyperparameters in $L$
\STATE Test $H$ with $E$, get weakness regions $\mathcal{R}$ 
\STATE Sample $m$ examples $\{f\}$ from $\mathcal{R}$,  $\mathcal{F}\leftarrow\mathcal{F} \cup \{f\}$
\STATE Fine-tune $H$ on $\mathcal{F}$ and $\mathcal{T}$ with $\epsilon$-sample for one epoch
\ENDFOR
\end{algorithmic} 
\end{algorithm}

\section{Limitations and Future Work}
\label{sec:lim}
\OURS finds the failure modes with a simulator. Our experimental results show that the current model can already find failure modes in the synthetic space that generalize well and fool models in real images, providing meaningful observations. We also demonstrate that these simulated images significantly improve the performance of current SOTA methods on real-world data. However, more realistic simulators are always helpful, but they may require additional rendering time. In the near future, we plan to explore more advanced simulators to further narrow down the domain gap with real data.

Currently, our work mainly focuses on one factor. Specifically, we study the robustness of HPS methods towards articulated pose, shape, and global rotation separately. However, when we optimize multiple factors at the same time, one factor usually dominates the others. For example, as previously mentioned, if we study the global rotation and the articulated pose at the same time, the global rotation will quickly converge to a maximum that has a very extreme viewpoint before the articulated pose converges, which makes the information we learn about the articulated pose less useful. To solve this issue, our current solution is to set limits to the global rotation. In the near future, we will study the solution for optimizing multiple non-independent and biased factors.

\end{document}